\documentclass[letterpaper]{article} 
\usepackage{aaai24}  
\usepackage{times}  
\usepackage{helvet}  
\usepackage{courier}  
\usepackage[hyphens]{url}  
\usepackage{graphicx} 
\usepackage{natbib}  
\usepackage{caption} 
\usepackage{newfloat}
\usepackage{listings}
\DeclareCaptionStyle{ruled}{labelfont=normalfont,labelsep=colon,strut=off} 
\usepackage{array}
\newcolumntype{C}[1]{>{\centering\let\newline\\\arraybackslash\hspace{0pt}}m{#1}}

\usepackage{color, colortbl}
\definecolor{Gray}{gray}{0.85}
\usepackage{lipsum}
\usepackage{kotex}
\usepackage{verbatim}
\usepackage{multirow}
\usepackage{amsmath}
\usepackage{multirow}
\usepackage{algpseudocode}
\usepackage{algorithm}
\usepackage{algorithmicx}
\usepackage{graphicx}
\usepackage{adjustbox}
\usepackage{comment}
\usepackage{helvet}
\usepackage{tikz}
\usetikzlibrary{fadings,shapes.arrows,shadows}   
\tikzset{arrowstyle/.style={draw= black,single arrow,minimum height=#1, single arrow,
single arrow head extend=.4cm,align=center }}
\usepackage{amsthm}
\usepackage{amssymb}

\newtheorem*{remark}{Remark}

\newtheorem{proposition}{Proposition}
\usepackage{bibentry}

\newtheorem{assumption}{Assumption}

\usepackage{subfigure}
\usepackage{multicol}

\usepackage[capitalize]{cleveref}

\usepackage{bibentry}

\title{Doubly Perturbed Task Free Continual Learning}
\author{
    Byung Hyun Lee$^1$, \ Min-hwan Oh$^{2}$, \ Se Young Chun$^{1,3, \dagger}$ 
}

\affiliations{
     \textsuperscript{\rm 1}Department of Electrical and Computer Engineering, Seoul National University\\
     \textsuperscript{\rm 2}Graduate School of Data Science, Seoul National University\\
     \textsuperscript{\rm 3}INMC \& IPAI, Seoul National University\\
    ldlqudgus756@snu.ac.kr, \ minoh@snu.ac.kr, \ sychun@snu.ac.kr
}

\begin{document}

\maketitle

\let\thefootnote\relax\footnotetext{$\dagger$ Corresponding author.}

\begin{abstract}
Task Free online Continual Learning (TF-CL) is a challenging problem where the model incrementally learns tasks without explicit task information. Although training with entire data from the past, present as well as \textit{future} is considered as the gold standard, naive approaches in TF-CL with the current samples may be conflicted with learning with samples in the future, leading to catastrophic forgetting and poor plasticity. Thus, a proactive consideration of  an unseen future sample in TF-CL becomes imperative. Motivated by this intuition, we propose a novel TF-CL framework considering future samples and show that injecting adversarial perturbations on both input data and decision-making is effective. Then, we propose a novel method named Doubly Perturbed Continual Learning (DPCL) to efficiently implement these input and decision-making perturbations. Specifically, for input perturbation, we propose an approximate perturbation method that injects noise into the input data as well as the feature vector and then interpolates the two perturbed samples. For decision-making process perturbation, we devise multiple stochastic classifiers. We also investigate a memory management scheme and learning rate scheduling reflecting our proposed double perturbations. We demonstrate that our proposed method outperforms the state-of-the-art baseline methods by large margins on various TF-CL benchmarks.
\end{abstract}

\section{Introduction}
Continual learning (CL) addresses the challenge of effectively learning tasks when training data arrives sequentially. 
A notorious drawback of deep neural networks in a continual learning is \textit{catastrophic forgetting}~\cite{mccloskey1989catastrophic}. As these networks learn new tasks, they often forget previously learned tasks, causing a decline in performance on those earlier tasks.
If, on the other hand, we restrict the update in the network parameters to counteract the catastrophic forgetting, the learning capacity for newer tasks can be hindered.
This dichotomy gives rise to what is known as the \textit{stability-plasticity dilemma}~\cite{carpenter1987art,mermillod2013stability}.
The solutions to overcome this challenge fall into three main strategies: regularization-based methods~\cite{kirkpatrick2017overcoming,jung2020continual,wang2021afec}, rehearsal-based methods~\cite{lopez2017gradient,shin2017continual,shmelkov2017incremental,chaudhry2018efficient,chaudhry2021using}, and architecture-based methods~\cite{mallya2018packnet,serra2018overcoming}.

In Task Free CL (TF-CL)~\cite{aljundi2019task}, the model incrementally learns classes in an online manner agnostic to the task shift, which is considered more realistic and practical, but more challenging setup than offline CL~\cite{koh2022online,zhangsimple}.
The dominant approach to relieve forgetting in TF-CL is memory-based approaches ~\cite{aljundi2019task,pourcel2022online}. They employ a small memory buffer to preserve a few past samples and replay them when training on a new task, but the restrictions on the available memory capacity highly degenerate performance on past tasks. 

Recently, several works suggested evolving the data distribution in memory by perturbing memory samples~\cite{wang2022improving, jin2021gradient}.
Meanwhile, flattening the weight loss landscape has also shown benefits in CL setups~\cite{cha2020cpr,deng2021flattening}. However, most of the prior CL works primarily concentrated on past samples, often overlooking future samples. Note that many CL studies use ``i.i.d. offline" as the oracle method of the best possible performance, not only for past and present data but also for future data. Therefore, incorporating unknown future samples in the CL model could be helpful in reducing forgetting and enhancing learning when training with real future samples.

In this work, we first demonstrate an upper bound for the TF-CL problem with unknown future samples, considering both adversarial input and weight perturbation, which has not been fully explained yet. Based on the observation, we propose a method, doubly perturbed continual learning (DPCL), addressing adversarial input perturbation with perturbed function interpolation and weight perturbation, specifically for classifier weights, through branched stochastic classifiers. Furthermore, we design a simple memory management strategy and adaptive learning rate scheduling induced by the perturbation. In experiments, our method significantly outperforms the existing rehearsal-based methods on various CL setups and benchmarks. Here is the summary of our contributions.
\begin{itemize}
\item We propose an optimization framework for TF-CL and show that it has an upper bound which considers the adversarial input and weight perturbations.
\item Our proposed method, DPCL, uses perturbed function interpolation and branching stochastic classifiers for input and weight changes with perturbation-based memory management and adaptive learning rate.
\item The proposed method outperforms baselines on various CL benchmarks and can be adapted to existing algorithms, consistently improving their performance.
\end{itemize}

\section{Related Works}
\subsection{Continual Learning (CL)}
CL seeks to retain prior knowledge while learning from sequential tasks exhibiting data distribution shifts. Most existing CL methods~\cite{lopez2017gradient, kirkpatrick2017overcoming, chaudhry2018riemannian, zenke2017continual, rolnick2019experience,yoon2018lifelong, mallya2018piggyback, hung2019compacting} primarily focus on the offline setting, where the learner can repeatedly access task samples during training without time constraints under the distinct task definitions separating task sequences.

\subsection{Task Free Continual Learning (TF-CL)}
TF-CL~\cite{aljundi2019task, jung2023new, pourcel2022online} addresses more general scenario where the model incrementally learns classes in an online manner and the data distribution change arbitrarily without explicit task information.
The majority of existing TF-CL approaches fall under rehearsal-based approaches~\cite{aljundi2019task, he2020task, wang2022improving}. They store a small number of samples from previous data stream and later replay them alongside new mini-batch data. 
Thus, we focus on rehearsal-based methods due to their simplicity and effectiveness.

Recently, DRO~\cite{wang2022improving} proposed to edit memory samples by adversarial input perturbation, making it gradually harder to be memorized.
\citeauthor{raghavan2021formalizing} (\citeyear{raghavan2021formalizing}) showed that the CL problem has an upper bound whose objective is to minimize with adversarial input perturbation, but it didn't fully consider the TF-CL setup.
Meanwhile,~\citeauthor{deng2021flattening} (\citeyear{deng2021flattening}) demonstrated the effectiveness of applying adversarial weight perturbation on training and memory management for CL.
To our best knowledge, it has not been investigated yet considering both input and weight perturbation simultaneously in TF-CL, and our work will propose a method that takes both into account.

\subsection{Input and Weight Perturbations}
Injecting input and weight perturbations into a standard training scheme is known to be effective for robustness and generalization by flattening the input and weight loss landscape.
It is well known that flat input loss landscape is correlated to the robustness of performance of a network to input perturbations.
In order to enhance robustness, adversarial training (AT) intentionally smooths out the input loss landscape by training on adversarially perturbed inputs. There are alternative approaches to flatten the loss landscape, through gradient regularization~\cite{lyu2015unified, ross2018improving}, curvature regularization~\cite{moosavi2019robustness}, and local linearity regularization ~\cite{qin2019adversarial}.
Meanwhile, multiple studies have demonstrated the correlation between the flat weight loss landscape and the standard generalization gap~\cite{keskarlarge, neyshabur2017exploring}. 
Especially, adversarial weight perturbation~\cite{wu2020adversarial} effectively improved both standard and robust generalization by combining it with AT or other variants.

\begin{figure*}[t]
\begin{center}
\centerline{\includegraphics[width=0.9\textwidth]{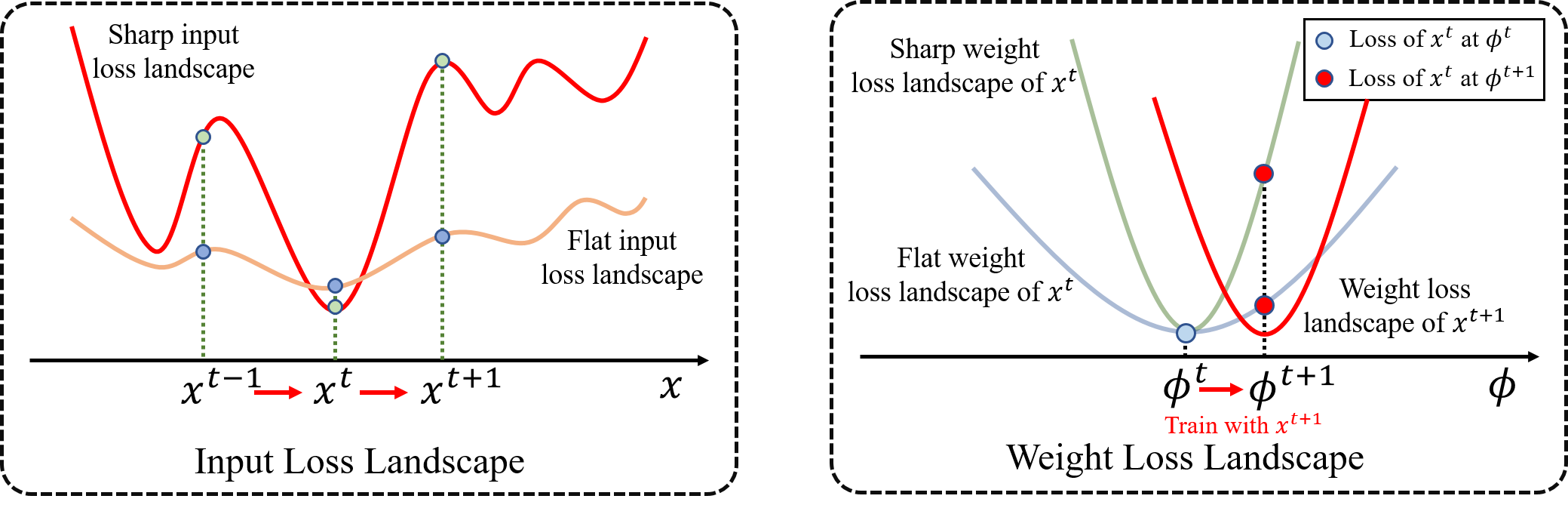}}
\caption{(Left) Input loss landscape of TF-CL when the weight $\theta^t$ has been determined for sample $x^t$.
We desire $\ell(h(x;\theta^t),y)$ to be flat about $x^t$ so that the loss for $x^{\tau}, \tau\in[1,\cdots,t-1, t+1]$ do not fluctuate significantly from $x^t$.
(Right) Weight loss landscape of TF-CL where $\phi$ gets shifted from $\phi^{t}$ to $\phi^{t+1}$ by training for new sample $x^{t+1}$.
We desire $\ell(h(x^t;[\theta_{e};\phi]),y^t)$ to be flat about $\phi^t$ so that the loss for $x^t$ doesn't increase dramatically when $\phi$ shifts from $\phi^t$ to $\phi^{t+1}$.}
\label{figure1}
\end{center}
\end{figure*}

\section{Problem Formulation}
\subsection{Revisiting Conventional TF-CL}
We denote a sample $(x,y) \in X \times Y$, where $X \subseteq \mathbb{R}^d$ is the input space (or image space), $Y \subseteq \mathbb{R}^C$ is the label space, and $C$ is the number of classes. A deep neural network to predict a label from an image can be defined as a function $h: X \to Y$, parameterized with $\theta$ and this learnable parameter $\theta$ can be trained by minimizing sample-wise loss $\ell (h(x; \theta), y)$. TF-CL is challenging due to varying data distribution $\mathcal{P}_t$ over iteration $t$, so the learner encounters a stream of data distribution $\{ \mathcal{P}_t\}_{t=0}^{T}$ via a stream of samples $\{(x^{t}, y^{t})\}_{t=1}^{T}$ where $(x^{t}, y^{t}) \sim \mathcal{P}_t$. Let us denote the sample-wise loss for $(x^{t}, y^{t})$ by $\mathcal{L}_{t}(\theta) = \ell (h(x^{t}; \theta), y^{t})$. Then, TF-CL trains the network $h$ in an online manner~\cite{aljundi2019gradient}:
\begin{equation}
   \theta^t \in \arg\min_\theta \mathcal{L}_{t}(\theta)
   \label{eq:oldtfcl}
\end{equation}
\[
\text{subject to}~\mathcal{L}_{\tau}(\theta) \le \mathcal{L}_{\tau}(\theta^{t-1}); \forall \tau \in [0,\ldots,t-1].
\]

\subsection{Novel TF-CL Considering a Future Sample}
Many CL studies regard the ``i.i.d. offline" training as the oracle due to its consistently low loss not only for past and present, but also for future data. Thus, considering the future samples could enhance the performance in TF-CL setup.
We first  relax the constraints in (\ref{eq:oldtfcl}) as a single constraint:
\begin{equation}
 \cfrac{1}{t}\textstyle\sum_{\tau = 0}^{t-1} \mathcal{L}_{\tau}(\theta) \le
 \cfrac{1}{t}\textstyle\sum_{\tau = 0}^{t-1} \mathcal{L}_{\tau}(\theta^{t-1}).
\label{eq:newtfcl1}
\end{equation}
Secondly, we introduce an additional constraint with a nuisance parameter $\theta'$ considering a future sample,
\begin{equation}
\mathcal{L}_{t+1}(\theta') \le \mathcal{L}_{t+1}(\theta).
\label{eq:newtfcl2}
\end{equation}
Then, using Lagrangian multipliers, the TF-CL with the minimization in (\ref{eq:oldtfcl}) with new constraints (\ref{eq:newtfcl1}) and (\ref{eq:newtfcl2}) will be
\begin{align}
     \theta^t \in \arg \min_{\theta} \,\, \mathcal{L}_t(\theta) & + \cfrac{\lambda}{t} \textstyle \sum_{\tau=0}^{t-1} (\mathcal{L}_{\tau}(\theta) - \mathcal{L}_{\tau}(\theta^{t-1})) \nonumber \\
     & + \rho (\mathcal{L}_{t+1}(\theta') - \mathcal{L}_{t+1}(\theta))
     \label{objective_proposed}
\end{align}
where $\lambda > 0$ and $\rho > 0$ are Lagrangian multipliers. 

\subsection{Doubly Perturbed Task Free Continual Learning}
Unfortunately, the future sample and most past samples are not available in TF-CL. Instead of minimizing the loss (\ref{objective_proposed}) directly, we minimize its surrogate independent of past and future samples. For this, we utilized the observation from \citeauthor{wu2019large} (\citeyear{wu2019large}) and \citeauthor{ahn2021ss} (\citeyear{ahn2021ss}) that change of parameter in classifier is more significant than change in encoder.

Let us consider the network $h = g \circ f$ that consists of the encoder $f$ and the classifier $g$, with the parametrization $h(\cdot;\theta) = g( f(\cdot; \theta_e); \phi)$ where $\theta = [\theta_e; \phi]$. Suppose that the new parameter $\theta' \approx [\theta_e; \phi']$ has almost no change in the encoder with the future sample $(x^{t+1},y^{t+1})$ while may have substantial change in the classifier. 
We also define $\eta_{1}^{t} := \max_{\tau} \lVert x^t-x^{\tau} \rVert < \infty, \tau=0,\ldots, t-1, t+1$ and $\eta_{2}^{t} := \max _{\phi'} \lVert \phi' - \phi^t \rVert$. Then, we have a surrogate of (\ref{objective_proposed}).

\begin{proposition}
Assume that $\mathcal{L}_t(\theta)$ is Lipschitz continuous for all $t$ and $\phi'$ is updated with finite gradient steps from $\phi^t$, so that $\phi'$ is a bounded random variable and $\eta_{2}^t < \infty$ with high probability. Then, the upper-bound for the loss (\ref{objective_proposed}) is
\begin{align}
     \mathcal{L} _t (\theta) & + \lambda \max _{\lVert \Delta x \rVert  \le \eta _{1}^{t} } \mathcal{L} _{t,\Delta}(\theta) \nonumber \\
     & + \rho \max _{\lVert \Delta \phi \rVert \le \eta _{2}^{t}} \max _{\lVert \Delta x \rVert \le \eta _{1}^{t}} \mathcal{L} _{t,\Delta}([\theta _e; \phi^t+ \Delta \phi]),
     \label{objective_upper_bound}
\end{align}
where $\mathcal{L}_{t,\Delta}(\theta) = \ell (h(x^t+\Delta x; \theta), y^t)$.
\label{proposition_surrogate}
\end{proposition}

Proposition \ref{proposition_surrogate} suggests that injecting adversarial perturbation on input and classifier's weight could help to minimize the TF-CL loss (\ref{objective_proposed}).
Note that both the second and third term have stably improved robustness and generalization of training~\cite{ross2018improving, wu2020adversarial}.
Here, $\eta_1^t$ handles the data distribution shift. For example, more intense perturbation is introduced for better robustness with large $\eta_1^t$ when crossing task boundaries.

Intuitively, such perturbations are known to find flat input and weight loss landscape~\cite{madry2017towards,foret2020sharpness}. 
For the input, it is desirable to achieve low losses for both past and future samples with the current network weights.
From Figure \ref{figure1}, a flatter input landscape is more conducive to achieving this goal. Moreover, if the loss of $x^t$ is flat about weights, then
one would expect only a minor increase in loss compared to a sharper weight landscape when the weights shift by training with new samples.
Since directly computing the adversarial perturbations is inefficient due to additional gradient steps, we approximately minimize this \emph{doubly perturbed} upper-bound (\ref{objective_upper_bound}) in an efficient way. 

\begin{figure*}[t]
\begin{center}
\centerline{\includegraphics[width=\textwidth]{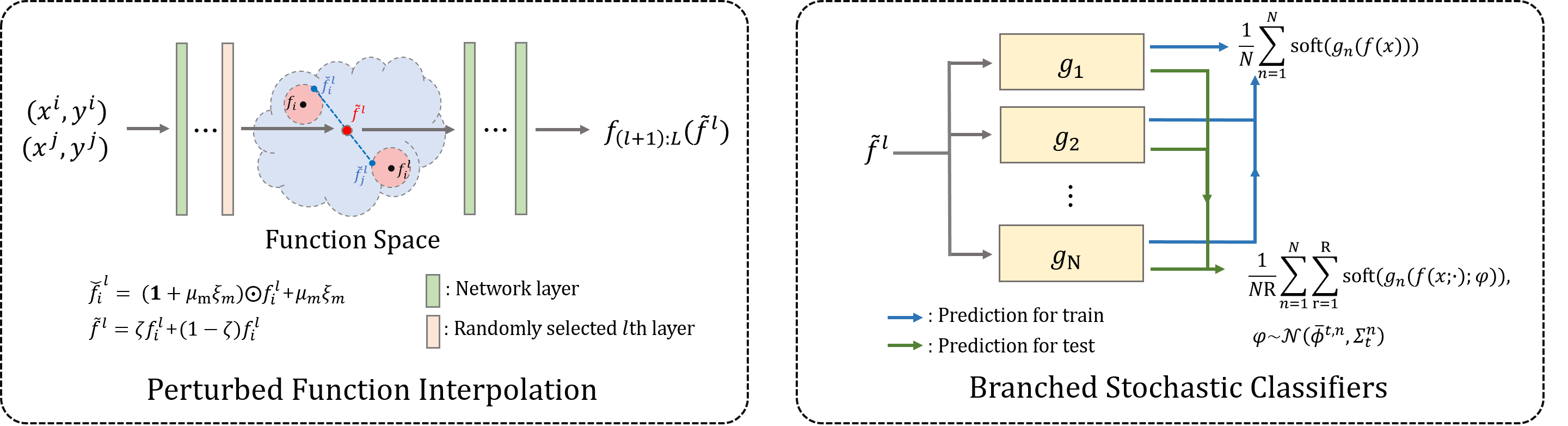}}
\caption{Illustration of Perturbed Function Interpolation (PFI) and Branched Stochastic Classifiers (BSC). PFI randomly perturbs the input, which makes the input loss landscape smooth. For weight perturbation, branched stochastic classifier utilizes weight average along the training trajectory, introduces multiple classifiers, and conduct variational inference during test.}
\label{fig2}
\end{center}
\end{figure*}

\section{Efficient Optimization for Doubly Perturbed Task Free Continual Learning}
In this section, we propose a novel CL method, called \textbf{D}oubly \textbf{P}erturbed \textbf{C}ontinual \textbf{L}earning (\textbf{DPCL}), which is inexpensive but very effective to handle the loss (\ref{objective_upper_bound}) with efficient input and weight perturbation schemes. We also design a simple memory management and adaptive learning rate scheme induced by these perturbation schemes. 

\subsection{Perturbed Function Interpolation}
\label{PFI}
Minimizing the second term in (\ref{objective_upper_bound}) requires gradient for input, which is heavy computation for online learning.
From~\citeauthor{limnoisy} (\citeyear{limnoisy}), we design a \textbf{P}erturbed \textbf{F}unction \textbf{I}nterpolation (\textbf{PFI}), a surrogate of the second term in (\ref{objective_upper_bound}).
Let the encoder $f$ consist of $L$-layered networks, denoted by $f=f_{(l+1):L} \circ f_{0:l}$, where $f_{0:l}$ maps an input to the hidden representation at $l$th layer (denoted by $f^l$) and $f_{(l+1):L} $ maps $f^l$ to the feature space of the encoder $f$. 
We define the average loss for samples whose label is $y$ as $\bar{\ell}_{y} = \sum_{\tau: y^\tau = y} \ell (h(x^\tau; \theta^\tau), y^\tau) / | \tau: y_\tau = y |$.  Then, for a randomly selected $l$th layer of $f$, the hidden representation of a sample is perturbed by noise considering its label:
\begin{equation}
    \check{f}^l = (\mathbf{1}+\mu_{m}\xi_{m}) \odot f^l + \mu_{a}\xi_{a}, \,\, \xi_{m}, \xi_{a} \sim \mathcal{N}(0, I),
    \label{noise_perturbation}
\end{equation}
where $\mathbf{1}$ denotes the one vector, $\odot$ is the Hadamard product, $I$ is the identity matrix, $\mu_m = \sigma_m \tan^{-1}(\bar{\ell}_{y})$, $\mu_a = \sigma_a \tan^{-1}(\bar{\ell}_{y})$, and $\sigma_m$, $\sigma_a$ are hyper-parameters. When label $y$ is first encountered, we set $\mu_m=\sigma_m$, $\mu_a=\sigma_a$. Instead of computing true $\bar{\ell}_y$, it is updated by exponential moving average whenever a sample of label $y$ is encountered.

As the main step, the function interpolation can be implemented for the two perturbed feature representations $\check{f}_i^l$ and $\check{f}_j^l$ with their labels $y^i$ and $y^j$, respectively, as follows:
\begin{equation}
 (\tilde{f}^l, \tilde{y}) = ( \zeta \check{f}_{i}^{l} + (1-\zeta) \check{f}_{j}^{l}, \zeta y_i + (1-\zeta) y_{j} ),
 \label{noise_perturbation_lin}
\end{equation}
where $\zeta \sim Beta(\alpha, \beta)$, $Beta(\cdot,\cdot)$ is the beta distribution. 
Finally, the output is computed by $f_{(l+1):L}(\tilde{f}^l)$ and we denote this as
$\tilde{f}(\cdot)$. Since PFI only requires element-wise multiplication and addition with samplings from simple distributions, its computational burden is negligible. 

\citeauthor{limnoisy} (\citeyear{limnoisy}) has shown that using perturbation and Mixup in function space can be interpreted as the upper bound of adversarial loss for the input under simplifying assumptions including that the task is binary classification. We extend it to multi-class classification setup assuming the classifier is linear for each class node and its node is trained in terms of binary classification.
Let $\tilde{\mathcal{L}}_\tau(\theta) = \ell(g(\tilde{f}(x^\tau)), y^\tau)$ is the loss computed by PFI.

\begin{proposition}
Assume that $\mathcal{L}_\tau(\theta)$ is computed by binary classifications for multi-classes.
We also suppose $\lVert \nabla_\theta h(x^\tau;\theta) \rVert >0$, $ d_1 \le \lVert f^l_{\tau} \rVert_2 \le d_2$ for some $0<d_1\le d_2$. With more regularity assumptions in~\citeauthor{limnoisy}(\citeyear{limnoisy}),
\begin{align}
\tilde{\mathcal{L}}_\tau(\theta) \ge \max_{\lVert \delta \rVert \le \epsilon } 
\ell(h(x^\tau+\delta; \theta), y^\tau) + \mathcal{L}_\tau^\mathrm{reg} + \epsilon^2 \psi_1(\epsilon),
\label{ineq_proposition}
\end{align}
where $\psi_1(\epsilon) \rightarrow 0$ and $\mathcal{L}_\tau^\mathrm{reg} \rightarrow 0$ as $\epsilon \rightarrow 0$, and
$\epsilon$ is assumed to be small and determined by each sample.
\label{proposition_pfi}
\end{proposition}
In Proposition \ref{proposition_pfi}, both $\mathcal{L}_\tau^{reg}$ and $\epsilon^2 \psi_1(\epsilon)$ are negligible for small $\epsilon$ and the adversarial loss term becomes dominant in the right side of (\ref{ineq_proposition}), which validates the use of PFI. See the supplementary materials for the details on Proposition \ref{proposition_pfi}.

\begin{table*}[t]
\centering
\normalsize
\setlength{\tabcolsep}{4pt}
\renewcommand{\arraystretch}{1.025}

\begin{tabular}{ C{4.75cm} C{1.75cm} C{1.75cm} C{1.75cm} C{1.75cm} C{1.75cm} C{1.75cm} }

 \hline
 \multirow{2}{*}{Methods}   & \multicolumn{2}{c}{CIFAR100 (M=2K)} & \multicolumn{2}{c}{CIFAR100-SC (M=2K)} & \multicolumn{2}{c}{ImageNet-100 (M=2K)} \\
                            \cline{2-7} 
                            & ACC$\uparrow$ & FM$\downarrow$& ACC$\uparrow$ & FM$\downarrow$ & ACC$\uparrow$ & FM$\downarrow$
 \\
 \hline
 
 ER ~\cite{rolnick2019experience}
 & 36.87$\pm$\small 1.53 & 44.98$\pm$\small 0.91
 & 40.09$\pm$\small 0.62 & 30.30$\pm$\small 0.60
 & 22.35$\pm$\small 0.29 & 51.87$\pm$\small 0.24
 \\

 EWC++ ~\cite{chaudhry2018riemannian}
 & 36.35$\pm$\small 1.62 & 44.23$\pm$\small 1.21
 & 39.87$\pm$\small 0.93 & 29.84$\pm$\small 1.04
 & 22.28$\pm$\small 0.45 & 51.50$\pm$\small 1.42
 \\

 DER++~\cite{buzzega2020dark}
 & 39.34$\pm$\small 1.01 & 40.97$\pm$\small 2.37
 & 41.54$\pm$\small 1.79 & 29.82$\pm$\small 2.26
 & 25.20$\pm$\small 2.06 & 52.16$\pm$\small 3.26
 \\

 BiC~\cite{wu2019large}
 & 36.64$\pm$\small 1.73 & 44.46$\pm$\small 1.24
 & 38.63$\pm$\small 1.32 & 29.96$\pm$\small 1.54
 & 22.41$\pm$\small 1.23 & 50.94$\pm$\small 1.34
 \\

 MIR~\cite{Aljundi2019OnlineCL}
 & 35.13$\pm$\small 1.35 & 45.97$\pm$\small 0.85
 & 37.84$\pm$\small 0.86 & 31.55$\pm$\small 1.00
 & 22.75$\pm$\small 1.03 & 52.65$\pm$\small 0.85
 \\
 
 CLIB~\cite{koh2022online}
 & 37.48$\pm$\small 1.27 & 42.66$\pm$\small 0.69
 & 37.27$\pm$\small 1.63 & 30.04$\pm$\small 1.85
 & 23.85$\pm$\small 1.36 & 49.96$\pm$\small 1.69
 \\

 ER-CPR~\cite{cha2020cpr}
 & 40.98$\pm$\small 0.12 & 44.47$\pm$\small 0.45
 & \underline{41.93$\pm$\small 0.42} & 30.67$\pm$\small 0.39
 & 27.08$\pm$\small 3.26 & 49.93$\pm$\small 1.06
 \\

 FS-DGPM~\cite{deng2021flattening}
 & 38.03$\pm$\small 0.58 & \underline{39.90$\pm$\small 0.39}
 & 37.03$\pm$\small 0.57 & 31.05$\pm$\small 1.63
 & 25.73$\pm$\small 1.68 & 49.32$\pm$\small 2.03
 \\

 DRO~\cite{ye2022task}
 & 39.23$\pm$\small 0.74 & 41.57$\pm$\small 0.25
 & 39.86$\pm$\small 0.95 & \underline{27.76$\pm$\small 0.77}
 & \underline{27.68$\pm$\small 1.23} & \underline{39.96$\pm$\small 0.87}
 \\

 ODDL~\cite{wang2022improving}
 & \underline{41.49$\pm$\small 1.38} & 40.01$\pm$\small 0.52
 & 40.82$\pm$\small 1.16 & 29.06$\pm$\small 1.87
 & 27.54$\pm$\small 0.63 & 41.23$\pm$\small 1.06
 \\

\hline

 DPCL  
 & \textbf{45.27$\pm$\small 1.32} & \textbf{37.66$\pm$\small 1.18}
 & \textbf{45.39$\pm$\small 1.34} & \textbf{26.57$\pm$\small 1.63}
 & \textbf{30.92$\pm$\small 1.17} & \textbf{37.33$\pm$\small 1.53}
\\
 \hline
\end{tabular}
\caption{Results on CIFAR100, CIFAR100-SC, and ImageNet-100. The tasks are distinguished by disjoint sets of classes. For all datasets, we measured averaged accuracy (ACC) and forgetting measure(FM) (\%) averaged by 5 different random seeds.}
\label{table_datasets}
\end{table*}


\subsection{Branched Stochastic Classifiers}
\label{BSC}
Minimizing the third term in (\ref{objective_upper_bound}) requires additional gradient steps. We bypass this inspired by ideas from~\citeauthor{izmailov2018swa} (\citeyear{izmailov2018swa}), ~\citeauthor{maddox2019simple} (\citeyear{maddox2019simple}), and~\citeauthor{wilson2020bayesian} (\citeyear{wilson2020bayesian}), updating multiple models by averaging their weights along training trajectories and performing variational inference for decisions of these models.~\citeauthor{cha2021swad} (\citeyear{cha2021swad}) confirmed that such stochastic weight averaging effectively flattens the weight loss landscape compared to the adversarial perturbation. In this work, we use this solely for the classifier, termed \textbf{B}ranched \textbf{S}tochastic \textbf{C}lassifiers (\textbf{BSC}).

We first consider a single linear classifier $g$, represented as $g = [g_1, \ldots g_C]$ where $g_c$ is the sub-classifier connected to the node for the $c$th class parameterized by $\phi_{c}$.
Considering the setup of TF-CL, we independently apply the weight perturbation to $g_c$.
Assume that $g_c$ at iteration $t$ follows a multivariate Gaussian distribution with mean $\bar{\phi}_{c}^t$ and covariance $\Sigma_t^c$. With the iteration $T_c$ when the model first encounters the class $c$, $\bar{\phi}_{c}^t$ and $\Sigma_t^c$ are updated every $P$ iterations:
\begin{align}
\bar{\phi}_{c}^{t} = \frac{k_c \bar{\phi}_{c}^{(t-P)} + \phi_c^{t}}{k_c+1},
\Sigma_{t}^{c} = \frac{1}{2}\Sigma_{diag,t}^c + \frac{D_{t,c}^T D_{t,c}}{2(A-1)},
\label{weight_avg_main}
\end{align}
where $k_c=\lfloor{(t-T_c)/P}\rfloor$ with floor function $\lfloor \cdot \rfloor$, $\Sigma_{diag,t}^c=diag( \overline{(\phi_{c}^{t})^{.2}} - (\bar{\phi}_{c}^{t})^{.2})$, $diag(v)$ is the diagonal matrix with diagonal $v$, and $(\cdot)^{.2}$ is the element-wise square. 
For (\ref{weight_avg_main}), 
$\overline{(\phi_{c}^{t})^{.2}}$ and $D_{t,c} \in \mathbb{R}^{q \times A}$ 
are updated as:
\begin{align}
\overline{(\phi_{c}^{t})^{.2}} & = \frac{k_c \overline{(\phi_{c}^{t-P})^{.2}} + (\phi_{c}^{t})^{.2} }{k_c+1}, \nonumber \\
D_{t,c} & = [D_{t-P,c}[2:A] \,\, (\phi_{c}^{t}-\bar{\phi}_{c}^{t})],
\label{weight_avg_requisite}
\end{align}
where $D[2:A]\in \mathbb{R}^{q \times (A-1)}$ is the submatrix of $D$ obtained by removing the first column of $D$. 
For inference, we predict the class probability $p^t$ using the variational inference with $\bar{\phi}^{t} = [\bar{\phi}_{1}^{t} \cdots \bar{\phi}_{C}^{t}]$ and $\Sigma_t = diag([\Sigma_{t,1} \cdots \Sigma_{t,C}])$ by 
\begin{equation}
p^t = \frac{1}{R} \sum_{r=1}^R \text{soft}(g(f(x;\theta_{e}^{t}); \varphi)), \quad \varphi \sim \mathcal{N}(\bar{\phi}^{t}, \Sigma_t),
\end{equation}
where $\text{soft}(\cdot)$ is the softmax function. We can view $\Sigma_t$ as the low-rank measures on deviation of the classifier parameters and the samplings as weight perturbation.

\citeauthor{wilson2020bayesian} (\citeyear{wilson2020bayesian}) verified that both deep ensembles and weight moving average can effectively improve the generalization performance. Since training multiple networks is time-consuming, we instead introduce multiple linear classifiers. With the decisions of the classifiers, the final decision for an input is determined by $\bar{p}^t = \frac{1} {N} \Sigma_{n=1}^N p^t_n$, where $N$ is the number of classifiers. In this case, each classifier can be viewed as an instance with perturbed weights.

\subsection{Perturbation-Induced Memory Management and Adaptive Learning Rate}
\label{PIMMLR}
Several studies showed the benefits of advanced memory management~\cite{chrysakis2020online} and adaptive learning rate~\cite{koh2022online} in TF-CL. To take the same advantage, we propose \textbf{P}erturbation-\textbf{I}nduced \textbf{M}emory management and \textbf{A}daptive learning rate (\textbf{PIMA}). For memory management, we basically balance the number of samples for each class in the memory $\mathcal{M}_t$ and compute the sample-wise mutual information empirically for a sample $(x,y)$ as 
\begin{equation}
\mathbb{I}(x;\phi^t) = \mathbb{H}(\bar{p}^t) - \frac{1}{N} \sum_{n=1}^{N} \mathbb{H}(p_n^t),
\end{equation}
where $\mathbb{H}(\cdot)$ is the entropy for class distribution.
To manage $\mathcal{M}_t$, we introduce a history $H_t$ which stores the mutual information for memory samples. Let $H_{t}(x, y)$ be the accumulated mutual information for a memory sample $(x, y)$ at $t$. If $(x,y)$ is selected for training, $H_t(x,y)$ is updated by 
\begin{equation}
 H_{t}(x, y) = (1-\gamma) H_{t-1}(x, y) + \gamma \mathbb{I}(x;\phi^t),
\end{equation}
where $\gamma \in (0,1)$. Otherwise, $H_{t}(x,y) = H_{t-1}(x,y)$.
To update the samples in the memory, we first identify the class $\hat{y}$ that occupies the most in $\mathcal{M}_t$. We then compare the values in $\{ H_t(x,y) | (x,y) \in \mathcal{M}_t, y=\hat{y} \}$ with $\mathbb{I}(x^{t};\phi^{t})$ for the current stream sample $(x^{t}, y^{t})$. If $\mathbb{I}(x^{t};\phi^{t})$ is the smallest, we skip updating the memory. Otherwise, we remove the memory sample of the lowest accumulated mutual information.

We also propose a heuristic but effective adaptive learning rate induced by $H_t$. Whenever $\bar{\phi}_{c}^{t}$ is updated, we scale the learning rate $\eta_t$ by a factor $\omega > 1$ if $\mathbb{E}_{(x,y) \sim \mathcal{M}_t} [ H_{t}(x, y)] > \mathbb{E}_{(x,y) \sim \mathcal{M}_t} [ H_{t-1}(x, y)]$ or $\frac{1}{\omega} < 1$ otherwise. The algorithm for the memory management and adaptive learning rate are explained in the supplementary materials.

\begin{figure*}[t]
\begin{center}
\centerline{\includegraphics[width=\textwidth]{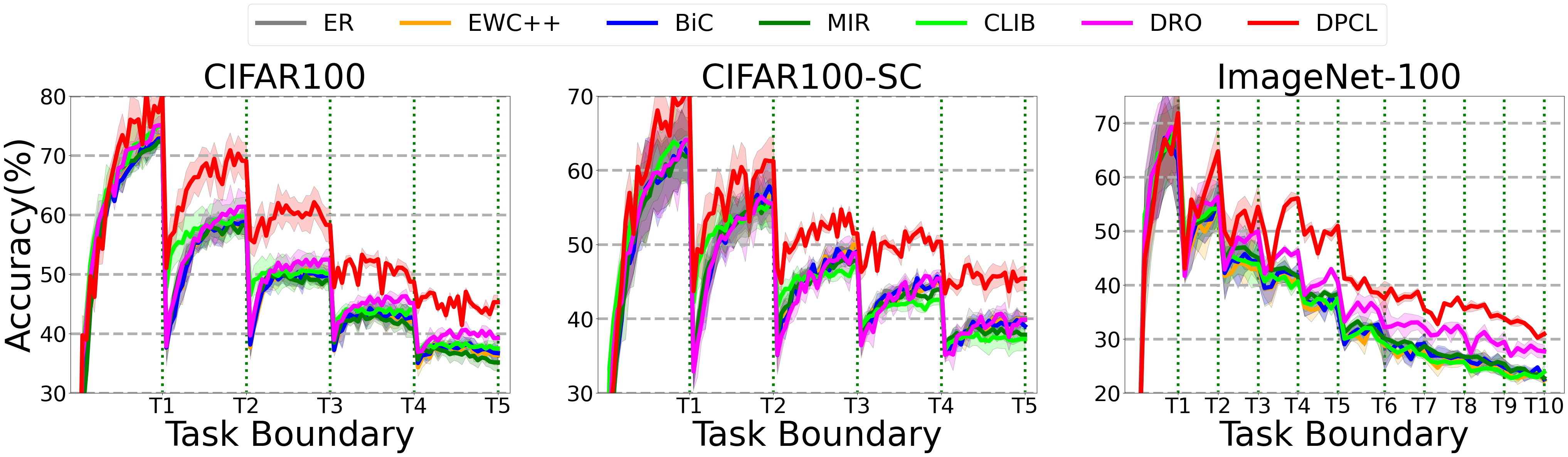}}
\caption{Any-time inference results on CIFAR100, CIFAR100-SC, and ImageNet-100. Each point represents average accuracy over 5 different random seeds and the shaded area represents the standard deviation($\pm$) around the average accuracy.
}
\label{figure_any^{t}ime_inference}
\end{center}
\end{figure*}

\section{Experiments}
\subsection{Experimental Setups}
\noindent\textbf{Benchmark datasets.}
We evaluate on three CL benchmark datasets. \textbf{CIFAR100}~\cite{rebuffi2017icarl} and \textbf{CIFAR100-SC}~\cite{yoon2019scalable} contains 50,000 samples and 10,000 samples for training and test. \textbf{ImageNet-100}~\cite{douillard2020podnet} is a subset of ILSVRC2012 with 100 randomly selected classes which consists of about 130K samples for training and 5000 samples for test. 
For both CIFAR100 and CIFAR100-SC, we split 100 classes into 5 tasks by randomly selecting 20 classes for each task~\cite{rebuffi2017icarl} and we considered the semantic similarity for CIFAR100-SC~\cite{yoon2019scalable}. For Imagenet-100, we split 100 classes into 10 tasks by randomly selecting 10 classes for each task~\cite{douillard2020podnet}.

\noindent\textbf{Task configurations.}
We also considered various setups: disjoint, blurry~\cite{bang2021rainbow}, and i-blurry~\cite{koh2022online} setups. \textbf{Disjoint} task configuration is the conventional CL setup where any two tasks don't share common classes. 
As a more general configuration, the \textbf{blurry} setup involves learning the same classes for all tasks while having different class distributions per task.
Meanwhile, each task in \textbf{i-blurry} setup consists of both shared and disjoint classes, which is more realistic than the disjoint and blurry setups.

\noindent\textbf{Baselines.}
Since most of TF-CL methods are rehearsal-based methods, we compared our DPCL with \textbf{ER}~\cite{rolnick2019experience}, \textbf{EWC++}~\cite{chaudhry2018riemannian}, \textbf{BiC}~\cite{wu2019large}, \textbf{DER++} \cite{buzzega2020dark}, and \textbf{MIR}~\cite{Aljundi2019OnlineCL}. By EWC++, we combined ER and the work of \citeauthor{chaudhry2018riemannian} (\citeyear{chaudhry2018riemannian}).
We compared DPCL with \textbf{CLIB}~\cite{koh2022online}, which was proposed for i-blurry CL setup. We also experimented \textbf{DRO}~\cite{wang2022improving}, which proposed to perturb the memory samples via distributionally robust optimization. Lastly, we experimented \textbf{FS-DGPM} \cite{deng2021flattening}, CPR \cite{cha2020cpr} by combining it with ER (\textbf{ER-CPR}), and \textbf{ODDL} \cite{ye2022task}.

\begin{table}[t]
\centering
\small
\setlength{\tabcolsep}{3pt}
\renewcommand{\arraystretch}{1.025}

\begin{tabular}{ C{1.4cm}  C{1.45cm} C{1.45cm} C{1.45cm}  C{1.45cm}}

 \hline
 \multirow{2}{*}{Methods}   & \multicolumn{2}{c}{Blurry (M=2K)} & \multicolumn{2}{c}{i-Blurry (M=2K)} \\
                            \cline{2-5} 
                            & ACC$\uparrow$ & FM$\downarrow$ & ACC$\uparrow$ & FM$\downarrow$ \\
 \hline
 
 ER
 & 24.24$\pm$\scriptsize 1.30 & 20.64$\pm$\scriptsize 2.50
 & 39.43$\pm$\scriptsize 1.09 & 15.45$\pm$\scriptsize 1.48
 \\

 EWC++
 & 23.84$\pm$\scriptsize 1.57 & 20.67$\pm$\scriptsize 3.34
 & 38.55$\pm$\scriptsize 0.79 & 15.57$\pm$\scriptsize 2.36
 \\

 DER++
 & 24.50$\pm$\scriptsize 3.03 & 17.35$\pm$\scriptsize 4.24
 & 44.34$\pm$\scriptsize 0.67 & 13.14$\pm$\scriptsize 4.64
 \\

 BiC
 & 24.96$\pm$\scriptsize 1.82 & 20.12$\pm$\scriptsize 3.78
 & 39.57$\pm$\scriptsize 0.90 & 14.23$\pm$\scriptsize 2.19
 \\

 MIR
 & 25.15$\pm$\scriptsize 0.08 & 15.49$\pm$\scriptsize 2.07
 & 38.26$\pm$\scriptsize 0.63 & 15.12$\pm$\scriptsize 2.69
 \\

 CLIB
 & \underline{38.13$\pm$\scriptsize 0.73} & \textbf{4.69$\pm$\scriptsize 0.99}
 & \underline{47.04$\pm$\scriptsize 0.89} & \underline{11.69$\pm$\scriptsize 2.12}
 \\

 ER-CPR
 & 28.72$\pm$\scriptsize 1.67 & 18.67$\pm$\scriptsize 1.23
 & 42.59$\pm$\scriptsize 0.66 & 18.01$\pm$\scriptsize 2.68
 \\

 FS-DGPM
 & 29.72$\pm$\scriptsize 0.22 & 14.51$\pm$\scriptsize 2.82
 & 41.99$\pm$\scriptsize 0.65 & 11.81$\pm$\scriptsize 0.12
 \\

 DRO
 & 20.86$\pm$\scriptsize 2.45 & 17.11$\pm$\scriptsize 2.47
 & 41.78$\pm$\scriptsize 0.42 & 11.97$\pm$\scriptsize 2.79
 \\

 ODDL
 & 33.35$\pm$\scriptsize 1.09 & 15.12$\pm$\scriptsize 1.98
 & 39.71$\pm$\scriptsize 1.32 & 16.12$\pm$\scriptsize 1.65
 \\

 \hline
 DPCL
 & \textbf{47.58$\pm$\scriptsize 2.75} & \underline{11.44$\pm$\scriptsize 2.64}
 & \textbf{50.22$\pm$\scriptsize 0.39} & \textbf{11.49$\pm$\scriptsize 2.54}
 \\
 \hline
\end{tabular}
\caption{Results on various setups on CIFAR100. We measured averaged accuracy (ACC) and forgetting measure (FM) (\%) averaged by 5 different seeds.}
\label{table_setups}
\end{table}


\noindent\textbf{Evaluation metrics.} We employ two primary evaluation metrics: averaged accuracy \textbf{(ACC)} and forgetting measure \textbf{(FM)}. ACC is a commonly used metric for evaluating CL methods~\cite{chaudhry2018riemannian, han2018co, van2018three}. 
FM is used to measure how much the average accuracy has dropped from the maximum value for a task ~\cite{yin2021mitigating, lin2022beyond}.
The details for the metric is explained in the supplementary materials. 

\noindent\textbf{Implementation details.} The overall experiment setting is based on \citeauthor{koh2022online} (\citeyear{koh2022online}). 
We used ResNet34~\cite{he2016deep} as the base feature encoder for all datasets. We used a batch size of 16 and 3 updates per sample for CIFAR100 and CIFAR100-SC and batch size of 64 and 0.25 updates per sample for ImageNet-100.
We used a memory size of 2000 for all datasets. We utilized the Adam optimizer~\cite{kingma2015adam} with an initial learning rate of 0.0003 and applied an exponential learning rate scheduler except CLIB and the optimization configurations reported in the original papers were used for CLIB. 
We applied AutoAugment \cite{cubuk2019autoaugment} and CutMix \cite{yun2019cutmix}. For DRO, we conducted the CutMix separately for the samples from stream buffer and memory since it conflicts with the perturbation for the memory samples. Since both utilizing CutMix and our PFI is ambiguous, we didn't apply CutMix for our method.
More information for the implementation details can be found in the supplementary materials.

\begin{figure*}[t]
\begin{center}
\centerline{\includegraphics[width=\textwidth]{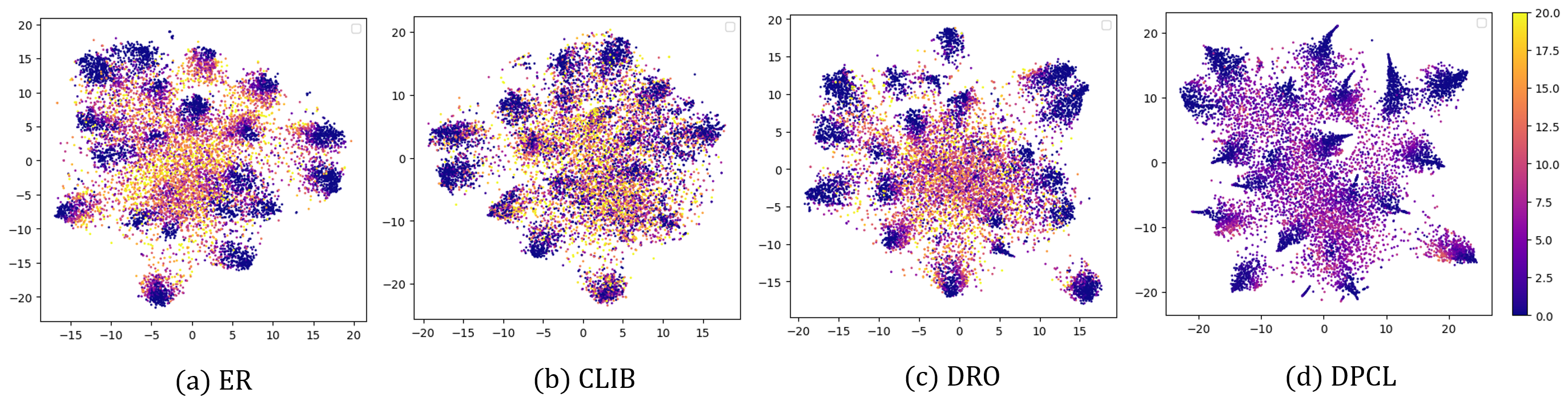}}
\caption{t-SNE on the features at the end of the encoder with CIFAR100. We computed the features and losses for samples in first task after training the last $5$th task. The color represents the loss of a sample (yellow for high loss and purple for low loss). We can see that our DPCL has overall low loss for all regions, especially near the class boundaries.}
\label{figure_analysis}
\end{center}
\end{figure*}

\begin{table}[t]
\centering
\small
\setlength{\tabcolsep}{2pt}
\renewcommand{\arraystretch}{1.05}

\begin{tabular}{ C{1.2cm} C{1.6cm} C{1.6cm} C{1.6cm} C{1.6cm} }
 \hline

 Methods & One-step (s)& Tr. Time (s) & Model Size & GPU Mem. \\
 \hline
 ER
 & 0.012 & 7126 & 1.00 & 1.00 
 \\
 EWC++
 & 0.027 & 16402 & 2.00 & 1.23 
 \\
 DER++
 & 0.019 & 11384 & 1.00 & 1.53 
 \\
 BiC
 & 0.015 & 10643 & 1.01 & 1.02 
 \\
 MIR
 & 0.029 & 18408 & 1.00 & 1.96 
 \\
 CLIB
 & 0.061 & 32519 & 1.00 & 4.32 
 \\
 ER-CPR
 & 0.015 & 8082 & 1.00 & 1.00 
 \\
 DRO
 & 0.038 & 23389 & 1.00 & 3.46 
 \\
 ODDL
 & 0.032 & 22905 & 2.14 & 3.31 
 \\
 \hline
 DPCL
 & 0.017 & 10925 & 1.03 & 1.06 
 \\
 \hline
\end{tabular}
\caption{Results of runtime/parametric complexity.
One-step (s): one-step throughput in second. Tr. Time (s): total training time in second. Model Size: normalized model size. GPU Mem.: normalized training GPU memory.}
\label{table_complexity}
\end{table}

\subsection{Main Results}
\noindent\textbf{Results on various benchmark datasets}
We first conducted experiments on CIFAR100, CIFAR100-SC, and ImageNet-100 in a disjoint setup. As shown in Table ~\ref{table_datasets}, DPCL significantly improves the performance on all three datasets. The extent of improvement of EWC++, BiC, and MIR was marginal compared to ER, as already observed in other studies~\cite{raghavan2021formalizing, ye2022task}. ER-CPR, FS-DGPM, DRO, and ODDL are perturbation-based methods and achieved  high performance on all datasets among baselines. Interestingly, we observed that for CIFAR100-SC, ER outperformed several baselines. CIFAR100-SC is divided into tasks based on their parent classes, and it seems that some existing advanced methods have limitations to improve upon ER's performance for the dataset in the challenging TF-CL scenario. On the other hand, our proposed method significantly outperformed ER for all datasets.
Recently,~\cite{koh2022online} argued that inference should be conducted at any-time for TF-CL since the model is agnostic to the task boundaries in practice. 
Based on this argument, we present the results of our method in terms of any-time inference in Fig. ~\ref{figure_any^{t}ime_inference}, where the vertical grids indicate task boundaries. From Fig. ~\ref{figure_any^{t}ime_inference}, our method consistently exhibits the best performance regardless of the iterations during training.


\noindent\textbf{Results on various task configurations.}
We evaluated our method on various task configurations. For this purpose, we evaluated the performance on CIFAR100 under the blurry ~\cite{bang2021rainbow} and i-blurry setup~\cite{koh2022online}. From Table \ref{table_setups}, we can observe that ER, EWC, BiC, and MiR show similar performance. CLIB, which is designed for the i-blurry setup, achieved the highest performance among the baselines. In contrast, DRO showed low performance in the blurry setup. On the other hand, our method consistently outperformed the baseline methods by a significant margin in both the blurry and i-blurry setups. Since there exists class imbalance in blurry or i-blurry settings, it empirically shows the robustness of the proposed method to class imbalance.

\noindent\textbf{Runtime/Parametric complexity analysis.}
In Table \ref{table_complexity}, we evaluated runtime and parametric complexity of the baselines and DPCL. One-step throughput (One-step) and total training time (Training Time) were measured in second for CIFAR100. We also measured model size (Model Size) and training GPU memory (GPU Mem.) on ImageNet-100 after normalizing their values for ER as 1.0. We didn't consider the memory consumption for replay memory since it is constant for all methods. From Table \ref{table_complexity}, we can see that our proposed method introduces mild increase on runtime, model size, and training GPU memory compared to other CL methods. Note that DRO, the gradient-based perturbation method, has significantly increased both the training time and memory consumption.

\begin{table}[t]
\hfill
\small
\centering
\footnotesize
\setlength{\tabcolsep}{2pt}
\renewcommand{\arraystretch}{1.025}

\begin{tabular}{ C{3.5cm} C{1.5cm} C{1.5cm}   }
    \hline
    \multirow{2}{*}{Methods}    & \multicolumn{2}{c}{CIFAR100 (Disjoint) } \\
                                \cline{2-3} 
                                & ACC$\uparrow$ & FM$\downarrow$ \\
    \hline
    
    DPCL w/o PFI
    & 40.85$\pm$\scriptsize 1.68 & 42.29$\pm$\scriptsize 2.32
    \\
    DPCL w/o BSC
    & 41.75$\pm$\scriptsize 1.43 & 40.56$\pm$\scriptsize 1.94
    \\
    DPCL w/o PIMA
    & 42.94$\pm$\scriptsize 1.23 & 39.79$\pm$\scriptsize 2.23
    \\
    DPCL
    & 45.27$\pm$\scriptsize 1.32 & 37.66$\pm$\scriptsize 1.18
    \\
        \hline
     BiC
     & 36.64$\pm$\scriptsize 1.73 & 44.46$\pm$\scriptsize 1.24
     \\
     BiC w/  PFI and BSC
    & 43.63$\pm$\scriptsize 0.74 & 38.57$\pm$\scriptsize 1.67
    \\
         \hline
     CLIB
     & 37.48$\pm$\scriptsize 1.27 & 42.66$\pm$\scriptsize 0.69
     \\
    CLIB w/ PFI and BSC
    & 44.97$\pm$\scriptsize 0.97 & 37.78$\pm$\scriptsize 1.24
    \\
    \hline
\end{tabular}
\caption{Ablation studies on CIFAR100. PFI and BSC were applied to BiC and CLIB. PIMA was excluded since it conflicts with the baseline methods.}
\label{table_ablation}
\end{table}

\noindent\textbf{Qualitative analysis.}
Fig. \ref{figure_analysis} shows the t-SNE results for samples within the first task 
after training the last task under the disjoint setup on CIFAR100. On the t-SNE map, we marked each samples with shade that represents the magnitude of the loss values for the corresponding features. From Fig. \ref{figure_analysis}, we can observe that our method produces much smoother loss landscape for features compared to baselines. It verifies that our method indeed flattens loss landscape in function space. For more experiments for the loss landscape, please refer to the supplementary materials.

\noindent\textbf{Ablation study.} To understand the effect of each component in our DPCL, we conducted an ablation study. We measured the performance of the proposed method by removing each component from DPCL. As shown in Table \ref{table_ablation}, we observed the obvious performance drop when removing each component, indicating the efficacy of each component of our method. Furthermore, to demonstrate the orthogonality of PFI and BSC to baselines, we applied them to other baselines such as BiC and CLIB. For this, we excluded the previously applied CutMix. Table \ref{table_ablation} confirms that the two components of our method can easily be combined with other methods to enhance performance.

\section{Conclusion}
In this work, we proposed a novel optimization framework for Task Free CL (TF-CL) and showed that it has an upper-bound which addresses the input and weight perturbations. Based on the framework, we proposed a method, Doubly Perturbed Continual Learning (DPCL), which employs perturbed function interpolation and incorporates branched stochastic classifiers for weight perturbations, with an upper-bound analysis considering adversarial perturbations. By additionally proposing simple scheme of memory management and adaptive learning rate, we could effectively improve the baseline methods on TF-CL.
Experimental results validated the superiority of DPCL over existing methods across various CL benchmarks and setups. 

\section*{Acknowledgments}

This work was supported in part by the National Research Foundation of Korea(NRF) grant funded by the Korea government(MSIT) (No. NRF-2022R1A4A1030579, 2022R1C1C100685912, NRF-2022M3C1A309202211), by Creative-Pioneering Researchers Program through Seoul National University, and by Institute of Information \& communications Technology Planning \& Evaluation (IITP) grant funded by the Korea government(MSIT) [NO.2021-0-01343, Artificial Intelligence Graduate School Program (Seoul National University)].

\bibliography{aaai24_main_cam_ready}

\clearpage
\appendix

\setcounter{figure}{0}
\setcounter{table}{0}
\setcounter{equation}{0}
\setcounter{proposition}{0}

\renewcommand{\thesection}{S\arabic{section}}
\renewcommand{\thefigure}{S\arabic{figure}}
\renewcommand{\thetable}{S\arabic{table}}
\renewcommand{\theequation}{S\arabic{equation}}

\onecolumn
\section{Doubly Perturbed Task-Free Continual Learning\\ \emph{Supplementary Materials}}
\subsection{}
\subsection{Algorithms for Doubly Perturbed Continual Learning (DPCL)}
In practice, a stream buffer is introduced for TF-CL which can store small number of stream samples. Furthermore, it is conventional to combine the stream buffer with multiple samples from the memory to construct the training batch for rehearsal-based approaches. We summarize the use of Perturbed Function Interpolation with the training batches.

\begin{minipage}[h]{0.95\textwidth}
    \begin{algorithm}[H]
    \small
    \caption{Perturbed Function Interpolation with training batches}
    \label{pfi}
    \begin{algorithmic}[1]
    
    \State \textbf{Input} training iteration $t$, 
    training batch $B_t$,
    $L$-layered encoder $f$, classifier $g$, mixing parameters $\alpha$ and $\beta$, multiplicative and additive noise factor $\sigma_m$ and $\sigma_a$, iteration encountered the label $y$ for the first time, $T_y$
    \\
    
    \State Select a random layer index $l \in \{0,1, \cdots, L \}$
    \State $B_{t}^{l} = \{ (f_{0:l}(x), y) | (x,y) \in B_{t}  \}$
    \State  $J_{t} = \{i | (f^l_i, y_i) \in B_t^l\}$ and $J^{'}_t = \{s_i \in J_t | i \in J_t\}$, where $J^{'}_t$ is the random permutation of $J_t$   
    \For{$(f^l_i,y_i) \in B_t^{l}$}
        \If{$T_{y_i} = t$}
            \State $\mu_m = \sigma_m$, $\mu_a = \sigma_a$
        \Else
            \State $\mu_m = \sigma_m  \text{tan}^{-1}(\ell_{avg}(y_i))$, $\mu_a = \sigma_a \text{tan}^{-1}(\ell_{avg}(y_i))$
        \EndIf
        \State $B_t^{l} \gets B_t^{l} \setminus \{(f^l_i, y_i)\} \cup \{((\mathbf{1}+\mu_{m}\xi_{m}) \odot f^{l}_{i} + \mu_{a}\xi_{a}, y_i)\}$, $\xi_m \sim \mathcal{N}(0, I)$ and $\xi_a \sim \mathcal{N}(0, I)$
    \EndFor
    
    \State $\tilde{B}_{t}^{l} = \{ ( \zeta f_{i}^{l} + (1-\zeta)f_{s_i}^{l}, \zeta y_i + (1-\zeta) y_{s_i} ) | (f_i^l, y_i) \in B_t^{l}, (f_{s_i}^l, y_{s_i})  \in B_t^{l} \} $, $\zeta \sim Beta(\alpha, \beta) $
    \State $F_{t} = \{ ( f_{(l+1):L}(\tilde{f}), \tilde{y} ) | (\tilde{f}^l, \tilde{y}) \in \tilde{B}_{t}^{l} \} $
    
    \State \textbf{Output} The network output with PFT, $F_t$
    \end{algorithmic}
    \label{algorithm-pfi}
    \end{algorithm}
    \vskip 0.2in
\end{minipage}

We also provide the algorithm for Perturbation-Induced Memory Management and Adaptive Learning Rate in the main paper through Algorithm \ref{pim} and Algorithm \ref{pia}. In similar to Algorithm \ref{algorithm-pfi}, we consider the stream buffer and training batch based on the memory usage of ER. 

\begin{multicols}{2}
    \begin{algorithm}[H]
    \small
    \caption{Perturbation-Induced Memory Management}
    \label{pim}
    \begin{algorithmic}[1]
    \State \textbf{Input} training iteration $t$, 
    stream buffer $S_t$, memory $\mathcal{M}$, memory budget size $m$, set of training memory samples $M_t$, importance history $H_t$
    \\
    \State $\gamma'=1-\gamma$
    \For{$(x^i, y^i)\in M_t$}
        \State $H_t(x^i,y^i) = \gamma'H_{t-1}(x^i,y^i) +\gamma \mathbb{I}(x^i;\phi^t)$
    \EndFor

    \For{$(x^\tau,y^\tau) \in S_t$}
        \If{$|\mathcal{M}| < m$}
            \State $\mathcal{M} \gets \mathcal{M}\cup \{ (x^t, y^t) \}$
        \Else
            \State $\bar{y} = \text{arg}\max_y | \{ (x^i, y^i) \in \mathcal{M} | y^i=y \} |$
            \State $\mathcal{I}_y = \{ j | (x^j, y^j) \in \mathcal{M}, y^j=\bar{y} \} $
            \State $\hat{j} = \text{arg}\min_{j\in\mathcal{I}_y} H_t(x^j, y^j)$
            \If{$\mathbb{I}(x^\tau;\phi^t) > H_t(x^{\hat{j}}, y^{\hat{j}})$}
                \State $\mathcal{M} \gets \mathcal{M} \backslash \{ (x^{\hat{j}}, y^{\hat{j}}) \} \cup \{ (x^{\tau}, y^{\tau}) \}$
            \EndIf
        \EndIf
    \EndFor

    \State \textbf{Output} $\mathcal{M}, H_t$
    \end{algorithmic}
    \end{algorithm}

    \columnbreak

    \begin{algorithm}[H]
    \small
    \caption{Perturbation-Induced Adaptive Learning Rate}
    \label{pia}

    \begin{algorithmic}[1]
    \State \textbf{Input} training iteration $t$, memory $\mathcal{M}$, weights of each sub-classifier by moving average $\bar{\phi}_c^t, c \in [1, \cdots, C]$, importance history $H_t$, learning rate $\eta_{t}$, multiplicative factor for learning rate schedule $\omega < 1$
    \\

    \For{$c \in [1, \cdots, C]$}
    
        \If{$\bar{\phi}_c^t \neq \bar{\phi}_c^{t-1}$}
            \If{$\mathbb{E}_{\mathcal{M}} [H_t] \ge \mathbb{E}_{\mathcal{M}} [H_{t-1}]$}
                \State $\eta_{t+1} = \omega \eta_{t}$
            \Else
                \State $\eta_{t+1} = \frac{\eta{t}}{\omega}$
            \EndIf
        \EndIf
        \State End the for loop
    \EndFor

    \State \textbf{Output} $\eta_{t+1}$
    \end{algorithmic}
   
    \end{algorithm}
\end{multicols}

\subsection{Proofs for Propositions}
The proposed TF-CL optimization problem in the main paper is
\begin{align}
     \theta^t \in \arg \min_{\theta} \,\, \mathcal{L}_t(\theta) + \cfrac{\lambda}{t} \textstyle \sum_{\tau=0}^{t-1} (\mathcal{L}_{\tau}(\theta) - \mathcal{L}_{\tau}(\theta^{t-1})) + \rho (\mathcal{L}_{t+1}(\theta') - \mathcal{L}_{t+1}(\theta)).
     \label{objective_proposed_supp}
\end{align}

Let us consider the network $h = g \circ f$ that consists of the encoder $f$ and the classifier $g$, with the parametrization $h(\cdot;\theta) = g( f(\cdot; \theta_e); \phi)$ where $\theta = [\theta_e; \phi]$. Suppose that the new parameter $\theta' \approx [\theta_e; \phi']$ has almost no change in the encoder with the future sample $(x^{t+1},y^{t+1})$ while may have substantial change in the classifier. 
We also define $\eta_{1}^{t} := \max_{\tau} \lVert x^t-x^{\tau} \rVert < \infty, \tau=0,\ldots, t-1, t+1$ and $\eta_{2}^{t} := \max _{\phi'} \lVert \phi' - \phi^t \rVert$. Then, we have a surrogate of (\ref{objective_proposed_supp}).

\begin{proposition}
Assume that $\mathcal{L}_t(\theta)$ is Lipschitz continuous for all $t$ and $\phi'$ is updated with finite gradient steps from $\phi^t$, so that $\phi'$ is a bounded random variable and $\eta_{2}^t < \infty$ with high probability. Then, the upper-bound for the loss (\ref{objective_proposed_supp}) is
\begin{align}
     \mathcal{L} _t (\theta) + \lambda \max _{\lVert \Delta x \rVert  \le \eta _{1}^{t} } \mathcal{L} _{t,\Delta}(\theta) + \rho \max _{\lVert \Delta \phi \rVert \le \eta _{2}^{t}} \max _{\lVert \Delta x \rVert \le \eta _{1}^{t}} \mathcal{L} _{t,\Delta}([\theta _e; \phi^t+ \Delta \phi]),
     \label{objective_upper_bound_supp}
\end{align}
where $\mathcal{L}_{t,\Delta}(\theta) = \ell (h(x^t+\Delta x; \theta), y^t)$.
\end{proposition}

\begin{proof}
Firstly, an equivalent loss to (\ref{objective_proposed_supp}) is 
\begin{equation}
     \mathcal{L}_t(\theta) + \cfrac{\lambda}{t} \textstyle \sum_{\tau=0}^{t-1} \mathcal{L}_{\tau}(\theta) + \rho (\mathcal{L}_{t+1}(\theta') - \mathcal{L}_{t+1}(\theta) )
      = \mathcal{L}_t(\theta) + \cfrac{\lambda}{t} \textstyle \sum_{\tau=0}^{t-1} \mathcal{L}_{\tau}(\theta) - \rho \mathcal{L}_{t+1}(\theta) + \rho \mathcal{L}_{t+1}(\theta').
\label{proposition1_rearange}
\end{equation}

Then, from the definition of $\eta_1^t$ and $\mathcal{L}_{\tau}(\theta) \le \max _{\lVert \Delta x \rVert  \le \eta _{1}^{t} } \mathcal{L} _{t,\Delta}(\theta) $ for all $\tau \in [0, \cdots, t+1]$, we have the bound
\begin{equation}
\cfrac{\lambda}{t} \textstyle \sum_{\tau=0}^{t-1} \mathcal{L}_{\tau}(\theta) - \rho \mathcal{L}_{t+1}(\theta) \le \lambda \max _{\lVert \Delta x \rVert  \le \eta _{1}^{t} } \mathcal{L} _{t,\Delta}(\theta), 
\label{proposition1_ineq_stability}
\end{equation}
where $\rho$ is assumed to be positive.

Without loss of generality, we assume that $\theta_e$ has no change while $\phi$ may change significantly. Then, we can find the upper-bound for $\rho \mathcal{L}_{t+1}(\theta')$ in (\ref{proposition1_rearange}) as
\begin{align}
\rho \mathcal{L}_{t+1}(\theta') \le \rho \max _{\lVert \Delta x \rVert  \le \eta _{1}^{t} } \mathcal{L} _{t,\Delta}([\theta_c;\phi']) \le \rho \max _{\lVert \Delta \phi \rVert \le \eta _{2}^{t}} \max _{\lVert \Delta x \rVert  \le \eta _{1}^{t} } \mathcal{L} _{t,\Delta}([\theta_c;\phi+\Delta \phi]).
\label{proposition1_ineq_plasiticity}
\end{align}
From (\ref{proposition1_rearange}), (\ref{proposition1_ineq_stability}), and (\ref{proposition1_ineq_plasiticity}), the upper-bound for the loss function of (\ref{objective_proposed_supp}) is
\begin{align}
     \mathcal{L}_t (\theta)
     + \lambda \max _{\lVert \Delta x \rVert  \le \eta _{1}^{t} } \mathcal{L} _{t,\Delta}(\theta)
     + \rho \max_{\lVert \Delta \phi \rVert \le \eta _{2}^{t}}  \max _{\lVert \Delta x \rVert  \le \eta _{1}^{t} } \mathcal{L} _{t,\Delta}([\theta_c;\phi+\Delta \phi]).
\end{align}
\end{proof}

\begin{remark}
Perturbed Feature Interpolation (PFI) is inspired from the Noisy Feature Mixup (NFM) ~\cite{limnoisy}. In our approach we inject perturbations on the features proportional to the loss values for each class and then use Mixup~\cite{zhang2017mixup}. On the other hand, NFM applies Mixup first and then perturbations are introduced to the features without considering the class. In other words, for two samples $(x^i,y^i)$ and $(x^j,y^j)$, the NFM for their features $(f^l_i,y^i)$ and $(f^l_j,y^j)$ can be expressed as follows:
\begin{align}
(\hat{f}^l, \tilde{y}) = ( (\mathbf{1}+\mu_{m}'\xi_{m}') \odot \text{Mix}_{\zeta'}(f^l_i, f^l_j) 
                       + \mu_{a}'\xi_{a}', \text{Mix}_{\zeta'}(y^i, y^j))
\end{align}
where $\text{Mix}_{\zeta'}(a,b) = \zeta'a+(1-\zeta')b$, $\xi_{m}'$ and $\xi_{a}' \sim \mathcal{N}(0, I)$, $\zeta' \sim Beta(\alpha, \beta)$, and $\mu_{m}'$ and $\mu_{a}'$ are hyper-parameters for noise scale.
Note that we can easily express PFI in the form of NFM by redefining the multiplicative noise scale $\mu_m$ and interpolation parameter $\zeta$, which proves that PFI and NFM are indeed equivalent.
\label{remark1_pfi_nfm_equiv}
\end{remark}

From \textbf{Remark}, we can verify that the properties of NFM can be utilized in same way for PFI. One remarkable theorem for NFM can be stated under some regularity conditions for a network~\cite{limnoisy}.
\begin{assumption}
Assume that the problem is the binary classification with sigmoid activation to compute the probability. Suppose that $\nabla_{f^l_{\tau}} h(x^\tau;\theta)$ and $\nabla_{f^l_{\tau}}^2 h(x^\tau;\theta)$ exist for all layers, $h(x^{\tau}) = \nabla_{f^{l}_{\tau}}h(x^{\tau})^{T}f^l_{\tau}$, $\nabla_{f^l_{\tau}}h(x^{\tau})=0$ for all $\tau$, and $\mathbb{E}_{\tau}[f^l_\tau] =0$, $\lVert \nabla_\theta h(x^\tau;\theta) \rVert >0$, $ d_1 \le \lVert f^l_{\tau} \rVert_2 \le d_2$ for some $0<d_1\le d_2$. 
\label{assumption}
\end{assumption}
Under the \textbf{Assumption ~\ref{assumption}}, ~\cite{limnoisy} has shown that the loss induced from the NFM is the upper-bound of the loss induced by adversarial perturbation in input space. By utilizing it, we state and prove the following proposition.

\begin{proposition}
Suppose that $\mathcal{L}_\tau(\theta) = \ell(h(x^\tau; \theta), y^\tau)$ is computed by binary classifications for multi-classes and $\tilde{\mathcal{L}}_\tau(\theta)$ is the loss computed with PFI. Also, assume that the classifier $g$ is linear for each class and can be represented as $g=[g_1, \cdots, g_C]$ for $C$ classes where $g_c$ is the sub-classifier connected to the node for the $c$th class, parameterized by $\phi_{c}$.
Then, under the \textbf{Assumption ~\ref{assumption}}, one can show that 
\begin{equation}
\tilde{\mathcal{L}}_\tau(\theta) \ge \max_{\lVert \delta \rVert \le \epsilon } 
\ell(h(x^\tau+\delta; \theta), y^\tau) + \mathcal{L}_\tau^\mathrm{reg} + \epsilon^2 \psi_1(\epsilon),
\label{ineq_proposition_supp}
\end{equation}
where 
$\psi_1$ is a function such that $\lim_{\epsilon\rightarrow 0}\psi_1(\epsilon) = 0$, $\epsilon$ is assumed to be small and determined depending on each sample and perturbations, $\mathcal{L}_\tau^\mathrm{reg} = \frac{1}{2C} \sum_{c=1}^C |S(g_c(f(x^\tau; \theta_e);\phi_c))|
(\epsilon^{reg})^2$ (detailed form for $\mathcal{L}_\tau^\mathrm{reg} = \frac{1}{2} |S(h(x^{\tau}))|(\epsilon^{reg})^2$ in the main text with the assumption of linear classifier $g$), $S(z) = \frac{e^{z}}{(1+e^z)^{2}}$, $(\epsilon^{reg})^2 = \epsilon^2 \lVert \nabla_{f^l} h(x^{\tau};\theta) \rVert_{2}^{2}\psi_2(\lambda, \theta, \sigma_{m}, \sigma_{a})$, and $\psi_2$ is bounded above.
\label{proposition_pfi_supp}
\end{proposition}

\begin{proof}
Since we consider the binary classification for each class in multi-class classification, we can represent the loss $\mathcal{L}_{\tau}(\theta)$ by the summation of class-wise losses as 
\begin{equation}
\mathcal{L}_{\tau}(\theta) = \ell(h(x^\tau; \theta), y^\tau) = \frac{1}{C}\sum_{c=1}^{C} \ell(g_c(f(x^\tau; \theta_e);\phi_c), y^\tau).
\end{equation}
In a similar way, we can express $\tilde{\mathcal{L}}_{\tau}(\theta)=\frac{1}{C}\sum_{c=1}^C \tilde{\mathcal{L}}_{\tau}^c(\theta)$ with the class-wise loss $\tilde{\mathcal{L}}_{\tau}^c(\theta)$ for perturbed features. From~\cite{limnoisy}, we can state an inequality for each class $c$ by 
\begin{equation}
\tilde{\mathcal{L}}^c_\tau(\theta) \ge \max_{\lVert \delta \rVert \le \epsilon } 
\ell(g_c(f(x^\tau+\delta; \theta_e);\phi_c), y^\tau) + \mathcal{L}_\tau^{c,\mathrm{reg}} + \epsilon^2 \psi_1^c(\epsilon),
\end{equation}
where 
$\psi_1^c$ is a function such that $\lim_{\epsilon\rightarrow 0}\psi_1^c(\epsilon) = 0$, $\epsilon$ is assumed to be small and determined depending on each sample and perturbations, $\mathcal{L}_\tau^{c,\mathrm{reg}} = \frac{1}{2} |S(g_c(f(x^\tau; \theta_e);\phi_c))|(\epsilon_c^{reg})^2$, $S(z) = \frac{e^{z}}{(1+e^z)^{2}}$, $(\epsilon_c^{reg})^2 = \epsilon^2 \lVert \nabla_{f^l} g_c(f(x^\tau; \theta_e);\phi_c) \rVert_{2}^{2}\psi_2^c(\lambda, \theta, \sigma_{m}, \sigma_{a})$, and $\psi_2^c$ is bounded above. Therefore, we have
\begin{align}
\tilde{\mathcal{L}}_\tau(\theta) \nonumber
& \ge \frac{1}{C}\sum_{c=1}^C \left( \max_{\lVert \delta \rVert \le \epsilon } 
\ell(g_c(f(x^\tau+\delta; \theta_e);\phi_c), y^\tau) + \mathcal{L}_\tau^{c,\mathrm{reg}} + \epsilon^2 \psi_1^c(\epsilon) \right) \\
        & \ge  \max_{\lVert \delta \rVert \le \epsilon } \frac{1}{C}\sum_{c=1}^C 
\ell(g_c(f(x^\tau+\delta; \theta_e);\phi_c), y^\tau) + \frac{1}{C}\sum_{c=1}^C \mathcal{L}_\tau^{c,\mathrm{reg}} + \epsilon^2 \frac{1}{C}\sum_{c=1}^C \psi_1^c(\epsilon) \\
        & \ge \max_{\lVert \delta \rVert \le \epsilon } 
\ell(h(x^\tau+\delta; \theta), y^\tau) + \mathcal{L}_\tau^\mathrm{reg} + \epsilon^2 \psi_1(\epsilon),
\end{align}
where the second inequality comes from the convexity of $\max$ function and the last inequality is derived by defining $\psi_2(\lambda,\theta,\sigma_m,\sigma_a) = \min_c \psi_2^c(\lambda, \theta, \sigma_m, \sigma_a)$
\end{proof}


\subsection{Experiment Details}
\noindent\textbf{Evaluation metrics.}
The two primary evaluation metrics in our work is the last average accuracy and forgetting measure. 
The average accuracy (\textbf{ACC}) can be computed by $A_{t}=\frac{1}{t}\sum_{i=1}^{t}a_{t,i}$ where $a_{t,i}$ is the accuracy for $i$-th task after training $t$-th task. It can measure the overall performance for tasks trained so far, but it is hard to measure the stability and plasticity.
By forgetting measure \textbf{(FM)}, we measure the drop of the average accuracy from the maximum value for a task ~\cite{yin2021mitigating, lin2022beyond}. Let $m_{i}^{t}$ be defined as $m_{i}^{t}=|a_{i,i}-a_{t,i}|$. Then, the average forgetting measure $F_{t}$ after training the $t$-th task is defined as $F_{t}=\frac{1}{t-1}\sum_{i=1}^{t-1}m_{i}^{t}$. \\

\noindent\textbf{The choice of CIFAR100-SC.} We chose CIFAR100-SC because it is known to be more challenging than CIFAR100. In CIFAR100-SC, the classes in CIFAR100 are grouped as superclasses to construct tasks. Since the superclasses are semantically different, the domain shift among tasks must be severer than random split of CIFAR100. We included this response in the supplementary material. \\

\noindent\textbf{Information for baselines.}
ER~\cite{rolnick2019experience} and EWC++~\cite{chaudhry2018riemannian} utilize a reservoir sampling for memory management, which involves randomly removing samples from memory to make room for new ones. Additionally, EWC++ incorporates a regularization term in the training loss to penalize the significance of weights, effectively constraining shift of weights. Due to the impracticality of herding selection~\cite{rebuffi2017icarl} in online CL since it requires access to all task data for computing class mean, we replaced the herding selection in BiC~\cite{wu2019large} with reservoir sampling, following a similar approach described in ~\cite{koh2022online}. 
MIR~\cite{Aljundi2019OnlineCL} enhances memory utilization by initially selecting a subset of memory that is larger than the size of the training batch. From this subset, samples are chosen based on the expected increase in loss if they were trained with streamed data. This process enables effective model updates.
FS-DGPM~\cite{deng2021flattening} first regulates the flatness of the weight loss landscape of past tasks and dynamically adjusts the gradient subspace for the past tasks to improve the plasticity for new task.
CLIB~\cite{koh2022online} is designed to maintain balance of number of samples per class in memory and exclusively utilizes memory for training purposes. A streamed sample can only be trained after it has been stored in memory.
ODDL~\cite{ye2022task} develops a framework that derives generalization bounds based on the discrepancy distance between the visited samples and the entire information accumulated during training. Inspired by this framework, it estimates the discrepancy between samples in the memory and proposes a new sample selection approach based on the discrepancy.
DRO~\cite{wang2022improving} introduces evolution framework for memory under TF-CL setup by dynamically evolving the memory data distribution that prevents overfitting and handles the high uncertainty in the memory. To achieve this, DRO evolves the memory using Wasserstein gradient flow for the probability measure. 

\subsection{Experiments with Different Number of Splits}
Since we fixed the number of splits as 5 and 10 for CIFAR100/CIFAR100-SC and ImageNet100 respectively in the main paper, we experimented with 10/20 splits for CIFAR100/CIFAR100-SC and 5/20 splits for ImageNet100 under the disjoint setup. Table \ref{table_diverse_splits} shows that the superiority of the proposed method with different number of splits.

\begin{table}[h]
\vskip -0.05in
\centering
\setlength{\tabcolsep}{1.5pt}
\renewcommand{\arraystretch}{0.8}

\begin{tabular}{ C{4.5cm}  C{1.75cm} C{1.75cm} C{1.75cm} C{1.75cm} C{1.75cm} C{1.75cm}}

 \hline
 ACC$\uparrow$ & \multicolumn{2}{c}{CIFAR100} & \multicolumn{2}{c}{CIFAR100-SC} & \multicolumn{2}{c}{ImageNet100} \\
 \hline
 Num. Splits   & 10 Splits & 20 Splits & 10 Splits & 20 Splits & 5 Splits & 20 Splits \\
 \hline

 ER ~\cite{rolnick2019experience}
 & 34.31$\pm$\small 1.02 & 31.09$\pm$\small 1.26
 & 36.73$\pm$\small 1.13 & 34.33$\pm$\small 1.62
 & 28.60$\pm$\small 1.38 & 20.49$\pm$\small 0.81
 \\

 DER++~\cite{buzzega2020dark}
 & 34.87$\pm$\small 1.67 & 33.36$\pm$\small 1.92
 & 36.10$\pm$\small 1.30 & 35.19$\pm$\small 1.31
 & 29.38$\pm$\small 0.74 & 18.42$\pm$\small 1.64
 \\
 
 CLIB~\cite{koh2022online}
 & 35.88$\pm$\small 1.23 & 32.27$\pm$\small 1.40
 & 36.62$\pm$\small 0.92 & 33.82$\pm$\small 1.02
 & 27.65$\pm$\small 0.77 & 19.51$\pm$\small 0.84
 \\

 ER-CPR~\cite{cha2020cpr}
 & 36.31$\pm$\small 0.54 & 33.79$\pm$\small 0.93
 & 37.82$\pm$\small 0.72 & 34.01$\pm$\small 2.58
 & 28.86$\pm$\small 0.85 & 20.70$\pm$\small 1.16
 \\

 FS-DGPM~\cite{deng2021flattening}
 & 35.50$\pm$\small 0.72 & 32.58$\pm$\small 0.82
 & 38.47$\pm$\small 0.84 & 35.22$\pm$\small 1.07
 & 31.51$\pm$\small 1.29 & 22.33$\pm$\small 0.73
 \\

 DRO~\cite{ye2022task}
 & 37.29$\pm$\small 0.82 & 35.88$\pm$\small 1.09
 & 38.81$\pm$\small 1.03 & \underline{36.86$\pm$\small 1.66}
 & 32.23$\pm$\small 1.40 & \underline{24.61$\pm$\small 1.33}
 \\

 ODDL~\cite{wang2022improving}
 & \underline{38.23$\pm$\small 1.17} & \underline{36.27$\pm$\scriptsize 1.76}
 & \underline{39.12$\pm$\small 1.64} & 36.39$\pm$\scriptsize 1.63
 & \underline{32.44$\pm$\small 1.57} & 23.71$\pm$\scriptsize 0.82
 \\

 \hline
 DPCL
 & \textbf{40.62$\pm$\small 1.39} & \textbf{38.09$\pm$\small 2.08}
 & \textbf{41.29$\pm$\small 1.66} & \textbf{38.62$\pm$\small 1.42}
 & \textbf{33.81$\pm$\small 1.03} & \textbf{26.33$\pm$\small 1.72}
 \\
 \hline
\end{tabular}

\vskip -0.05in
\caption{Accuracies for disjoint setup on CIFAR100, CIFAR100-SC, and ImageNet100 with the various number of splits, averaged by 3 different seeds.} 
\label{table_diverse_splits}
\vskip -0.25in
\end{table}

\subsection{Experiments with Different Blurriness Parameters}
With $N_b$ and $M_b$ being the portion of disjoint classes and the portion of samples of minor classes in a task respectively, we fixed $(N_b, M_b)$ as $(0\%, 10\%)$ for blurry and $(50\%, 10\%)$ for i-blurry setups in Table 2 in the main paper. We explored different values for $M_b$ and $N_b$ in Table \ref{table_blurriness} and the proposed method still outperforms the baselines.

\begin{table}[h]
\vskip -0.0in
\centering
\setlength{\tabcolsep}{2pt}
\renewcommand{\arraystretch}{0.8}

\begin{tabular}{ C{1.8cm}  C{1.8cm} C{1.8cm} C{1.8cm}  C{1.8cm}}
 \hline
 ACC$\uparrow$   & \multicolumn{2}{c}{Blurry} & \multicolumn{2}{c}{i-Blurry} \\
 \hline
 $(N_b, M_b)$ & $(0\%, 20\%)$ & $(0\%, 30\%)$ & $(25\%, 10\%)$ & $(75\%, 10\%)$ \\
 \hline
 
 ER
 & 14.28$\pm$\small 1.31 & 22.83$\pm$\small 1.13
 & 37.96$\pm$\small 1.03 & 36.93$\pm$\small 0.78
 \\

 DER++
 & 17.32$\pm$\small 1.04 & 28.34$\pm$\small 0.84
 & 39.02$\pm$\small 0.98 & 39.60$\pm$\small 1.01
 \\
 CLIB
 & \underline{29.02$\pm$\small 0.67} & \underline{34.87$\pm$\small 0.71}
 & \underline{45.23$\pm$\small 1.14} & \underline{46.02$\pm$\small 0.90}
 \\

 ER-CPR
 & 13.05$\pm$\small 0.89 & 25.59$\pm$\small 0.87
 & 36.77$\pm$\small 0.82 & 36.97$\pm$\small 0.73
 \\

 FS-DGPM
 & 20.43$\pm$\small 0.83 & 31.81$\pm$\small 1.12
 & 39.92$\pm$\small 0.87 & 40.45$\pm$\small 0.94
 \\
 
 DRO
 & 13.40$\pm$\small 1.72 & 23.43$\pm$\small 1.94
 & 40.65$\pm$\small 1.16 & 39.96$\pm$\small 0.89
 \\

 ODDL
 & 26.04$\pm$\small 1.45 & 33.28$\pm$\small 1.67
 & 38.65$\pm$\small 0.76 & 40.13$\pm$\small 0.95
 \\

 \hline
 DPCL
 & \textbf{35.14$\pm$\small 1.85} & \textbf{43.99$\pm$\small 1.72}
 & \textbf{47.96$\pm$\small 1.07} & \textbf{47.33$\pm$\small 0.85}
 \\
 \hline
\end{tabular}

\vskip -0.05in
\caption{Accuracies for blurry/i-blurry setups on CIFAR100 with diverse blurriness parameters, averaged by 3 different seeds.} 
\label{table_blurriness}
\vskip -0.2in
\end{table}

\subsection{Ablation Studies on Hyper-parameters}
The proposed method has some hyper-parameters such as $\alpha$, $\beta$, $\sigma_a$, $\sigma_m$, and $N$.
For all experiments, we fixed $\alpha$$=$$\beta$$=$$1.0$ following~\citeauthor{limnoisy}(\citeyear{limnoisy}) and we have searched for values of $(\sigma_a,\sigma_m)$ on CIFAR100 and set $(0.4,0.2)$. Considering trade-off between computation and performance, we set $N=5$. To explore the effect of different values of the hyper-parameters ($\sigma_a$, $\sigma_m$, $N$), we provide Table \ref{table_hyperparams}, which shows that the proposed method is not sensitive to those hyper-parameters.
\begin{table}[h]
\vskip 0.05in
\centering
\setlength{\tabcolsep}{3pt}
\renewcommand{\arraystretch}{0.8}

\begin{tabular}{C{0.75cm} C{2.0cm} | C{0.75cm}  C{2.0cm} | C{0.75cm} C{2.0cm}}

 \hline

   $\sigma_a $ & ACC$\uparrow$ & $\sigma_m$ & ACC$\uparrow$ & $N$ & ACC$\uparrow$ \\
 \hline
 
   0.1 & 43.33$\pm$\small 1.28
 & 0.05 & 44.06$\pm$\small 1.06
 & 1    & 43.85$\pm$\small 1.90
 \\
 
   0.2  & 44.88$\pm$\small 1.44
 & 0.1  & 44.73$\pm$\small 1.61
 & 2    & 44.02$\pm$\small 2.13
 \\   
 
   0.4  & 45.27$\pm$\small 1.32
 & 0.2  & 45.27$\pm$\small 1.32
 & 5    & 45.27$\pm$\small 1.32
 \\
 
   0.8  & 42.97$\pm$\small 1.98
 & 0.4  & 44.23$\pm$\small 1.30
 & 10   & 45.79$\pm$\small 0.51
 \\

   1.6    & 41.20$\pm$\small 1.85
 & 0.8    & 42.39$\pm$\small 1.64
 & 20   & 46.02$\pm$\small 0.93
 \\
 \hline
\end{tabular}

\vskip -0.05in

\caption{Accuracies with different values of hyper-parameters for disjoint setup on CIFAR100 averaged by 3 different seeds.}
\label{table_hyperparams}
\vskip -0.0in
\end{table}

\subsection{Experiments with ResNet18}
Following the setup of~\citeauthor{koh2022online}(\citeyear{{koh2022online}}), we used ResNet34 for disjoint, blurry, and i-blurry setups in the main paper. Since ResNet18 is also a frequently used architecture, we evaluated our method with ResNet18 on CIFAR100, CIFAR100-SC under disjoint setup and reported the results in Table \ref{table_resnet18}, which shows that the proposed method still outperforms the others.

\begin{table}[h]
\vskip -0.0in
\centering
\setlength{\tabcolsep}{4pt}
\renewcommand{\arraystretch}{0.8}
\begin{tabular}{ C{1.75cm}  C{1.75cm} C{1.75cm} C{1.75cm}  C{1.75cm}}

 \hline
 \multirow{2}{*}{Methods}   & \multicolumn{2}{c}{CIFAR100} & \multicolumn{2}{c}{CIFAR100-SC} \\
                            \cline{2-5} 
                            & ACC$\uparrow$ & FM$\downarrow$ & ACC$\uparrow$ & FM$\downarrow$ \\
 \hline
 
 ER
 & 35.82$\pm$\small 1.54 & 43.23$\pm$\small 1.96
 & 37.42$\pm$\small 1.21 & 32.63$\pm$\small 1.62
 \\

 DER++
 & 39.01$\pm$\small 1.06 & 40.12$\pm$\small 2.01
 & 40.99$\pm$\small 0.77 & 29.18$\pm$\small 1.83
 \\
 CLIB
 & 36.45$\pm$\small 1.30 & 39.39$\pm$\small 0.95
 & 38.33$\pm$\small 1.09 & 29.52$\pm$\small 1.22
 \\

 ER-CPR
 & 37.09$\pm$\small 1.29 & 42.03$\pm$\small 2.68
 & 38.12$\pm$\small 1.16 & 31.55$\pm$\small 1.45
 \\

 FS-DGPM
 & 37.66$\pm$\small 1.44 & 39.95$\pm$\small 2.02
 & 38.58$\pm$\small 0.87 & 30.17$\pm$\small 1.04
 \\
 
 DRO
 & 38.97$\pm$\small 0.88 & \underline{38.12$\pm$\small 1.53}
 & 39.00$\pm$\small 1.01 & \underline{28.29$\pm$\small 1.70}
 \\

 ODDL
 & \underline{40.48$\pm$\small 1.92} & 38.87$\pm$\small 2.08
 & \underline{41.01$\pm$\small 1.21} & 28.45$\pm$\small 1.63
 \\

 \hline
 DPCL
 & \textbf{43.51$\pm$\small 1.71} & \textbf{37.19$\pm$\small 2.74}
 & \textbf{44.42$\pm$\small 1.41} & \textbf{27.55$\pm$\small 1.92}
 \\
 \hline
\end{tabular}

\vskip -0.0in
\caption{Results with ResNet18 under disjoint setup on CIFAR100 and CIFAR100-SC averaged by 3 different random seeds.} 
\label{table_resnet18}
\vskip -0.2in
\end{table}

\subsection{Comparison Experiments of PIMA}
Several studies have demonstrated that advanced memory management strategy can improve the performance~\cite{koh2022online, bang2021rainbow, chrysakis2020online}. Additionally, adjusting the learning rate appropriately also enhanced the performance in TF-CL~\cite{koh2022online}. We proposed PIMA to take the same advantages of them in our proposed doubly perturbed scheme leveraging the network output obtained through PFI and BSC with neglegible computation and memory consumption. 

To demonstrate its effectiveness experimentally, we conducted comparison experiments by replacing each element with other baselines. The below Table \ref{table_mem} and Table \ref{table_adl} present the results using other memory management and adaptive learning strategies on disjoint setup for CIFAR100, maintaining the other components of the proposed method, which confirm the efficacy of our PIMA. We added the tables in the supplementary material.

\begin{table}[h]
\centering
\normalsize
\setlength{\tabcolsep}{1pt}
\renewcommand{\arraystretch}{1.05}

\begin{tabular}{ C{1.4cm} C{1.8cm} C{1.8cm} C{1.8cm} C{1.8cm}  C{1.8cm}}
 \hline

 Methods & CLIB & RM & CBRS & Reservoir & DPCL \\
 \hline
 ACC$\uparrow$
 & 41.13$\pm$\normalsize 0.75 & 42.24$\pm$\normalsize 0.22 & 42.24$\pm$\normalsize 0.22 & 41.75$\pm$\normalsize 0.78 & 45.27$\pm$\normalsize 1.32 
 \\
 
 FM$\downarrow$ 
 & 38.59$\pm$\normalsize 2.08 & 38.20$\pm$\normalsize 2.43 & 38.84$\pm$\normalsize 0.75 & 37.91$\pm$\normalsize 1.55 & 37.66$\pm$\normalsize 1.18 
 \\
 \hline
\end{tabular}
\vskip -0.05in
\caption{Comparison of memory management schemes on disjoint setup on CIFAR100. We measured averaged accuracy (ACC.) and forgetting measure (FM.) (\%) averaged by 5 different seeds.}
\vskip -0.2in
\label{table_mem}
\end{table}

\begin{table}[h]
\centering
\normalsize
\setlength{\tabcolsep}{1pt}
\renewcommand{\arraystretch}{1.05}
\vskip -0.05in

\begin{tabular}{ C{1.4cm} C{1.8cm} C{1.8cm} }
 \hline

 Methods & CLIB & DPCL \\
 \hline
 ACC$\uparrow$ & 43.80$\pm$\normalsize 0.34 & 45.27$\pm$\normalsize 1.32 
 \\
 FM$\downarrow$  & 37.39$\pm$\normalsize 0.50 & 37.66$\pm$\normalsize 1.18 
 \\
 \hline
\end{tabular}
\vskip -0.05in
\caption{Comparison of adaptive learning rate schemes on disjoint setup on CIFAR100. We measured averaged accuracy (ACC.) and forgetting measure (FM.) (\%) averaged by 5 different seeds.}
\label{table_adl}
\vskip -0.15in
\end{table}

\subsection{Comparison Experiments under the Setup of DRO}
In order to demonstrate the effectiveness of our method in different settings, we conducted experiments under the setup of DRO~\cite{wang2022improving}, especially for comparison with DRO since it is one of the state-of-the-art (SOTA) methods in TF-CL. For this comparison, we conducted the experiments on CIFAR10 and on CIFAR100.

Following of work of \citeauthor{wang2022improving} (\citeyear{wang2022improving}), we split CIFAR10 into 5 tasks, set the memory size as 500, and omitted CutMix. For CIFAR100, we split it into 20 tasks
and evaluated under various memory sizes (1K, 2K, and 5K) and omitted CutMix. From Table \ref{table_dro}, we can see that DPCL consistently outperforms DRO both on CIFAR10 and CIFAR100.

\begin{table}[H]
\centering
\setlength{\tabcolsep}{1.2pt}
\renewcommand{\arraystretch}{1.2}
\vskip -0.0in
\begin{tabular}{ C{1.5cm} C{1.8cm} C{1.8cm} C{1.8cm} C{1.8cm} C{1.8cm} C{1.8cm} C{1.8cm} C{1.8cm}}
 \hline
 \multirow{2}{*}{Methods}   & \multicolumn{2}{c}{CIFAR10 (M=500)} & \multicolumn{2}{c}{CIFAR100 (M=1K)} & \multicolumn{2}{c}{CIFAR100 (M=2K)} & \multicolumn{2}{c}{CIFAR100 (M=5K)} \\
                            \cline{2-9} 
                            & DRO & DPCL & DRO & DPCL & DRO & DPCL & DRO & DPCL
 \\
 \hline
 
 ACC$\uparrow$ 
 & 50.03$\pm$\small 1.41 & 55.56$\pm$\small 0.75 
 & 18.37$\pm$\small 1.13 & 24.73$\pm$\small 1.32  
 & 27.42$\pm$\small 0.93 & 27.42$\pm$\small 0.93
 & 24.44$\pm$\small 1.13 & 29.73$\pm$\small 1.85
 \\
 FM$\downarrow$ 
 & 36.22$\pm$\small 2.03 & 32.93$\pm$\small 1.79
 & 34.32$\pm$\small 1.65 & 29.29$\pm$\small 1.82
 & 32.42$\pm$\small 2.76 & 27.55$\pm$\small 1.35
 & 29.31$\pm$\small 2.31 & 24.08$\pm$\small 1.94
 \\
 \hline
\end{tabular}
\vskip -0.0in
\caption{Comparison with DRO on disjoint setup on CIFAR10 and CIFAR100 with various memory size. We measured averaged accuracy (ACC.) and forgetting measure (FM.) (\%) averaged by 5 different seeds.}
\vskip -0.0in
\label{table_dro}
\end{table}

\subsection{Comparison to Architecture-based Methods.}
We conducted additional experiments for architecture-based methods on TF-CL, reported in Table \ref{table_architecture_datasets} and \ref{table_architecture_setups}. We compared our method with CTN~\cite{pham2020contextual}, evaluated for online CL, and CCLL~\cite{singh2020calibrating}, evaluated only for offline CL. To our best knowledge, there's no architecture-based method for TF-CL, so we conducted them under task-aware setting. The results demonstrate that the proposed method consistently outperforms CTN and CCLL.

\begin{table*}[h]
\centering
\normalsize
\setlength{\tabcolsep}{4pt}
\renewcommand{\arraystretch}{1.025}

\begin{tabular}{ C{4.75cm} C{1.75cm} C{1.75cm} C{1.75cm} C{1.75cm} C{1.75cm} C{1.75cm} }

 \hline
 \multirow{2}{*}{Methods}   & \multicolumn{2}{c}{CIFAR100 (M=2K)} & \multicolumn{2}{c}{CIFAR100-SC (M=2K)} & \multicolumn{2}{c}{ImageNet-100 (M=2K)} \\
                            \cline{2-7} 
                            & ACC$\uparrow$ & FM$\downarrow$& ACC$\uparrow$ & FM$\downarrow$ & ACC$\uparrow$ & FM$\downarrow$
 \\
 \hline
 
 CTN~\cite{pham2020contextual}
 & 32.56$\pm$\small 1.76 & 46.12$\pm$\small 1.93
 & 29.12$\pm$\small 1.56 & 36.43$\pm$\small 2.62
 & 18.07$\pm$\small 2.08 & 55.21$\pm$\small 2.82
 \\
 CCLL~\cite{singh2020calibrating}
 & 29.32$\pm$\small 1.37 & 47.12$\pm$\small 1.93
 & 27.12$\pm$\small 1.75 & 37.12$\pm$\small 0.89
 & 17.73$\pm$\small 1.31 & 54.57$\pm$\small 1.98
 \\

\hline

 DPCL  
 & \textbf{45.27$\pm$\small 1.32} & \textbf{37.66$\pm$\small 1.18}
 & \textbf{45.39$\pm$\small 1.34} & \textbf{26.57$\pm$\small 1.63}
 & \textbf{30.92$\pm$\small 1.17} & \textbf{37.33$\pm$\small 1.53}
\\
 \hline
\end{tabular}
\caption{Results of architecture-based methods on CIFAR100, CIFAR100-SC, and ImageNet-100 on disjoint setup. For all datasets, we measured averaged accuracy (ACC) and forgetting measure(FM) (\%) averaged by 5 different random seeds.}
\vskip -0.1in
\label{table_architecture_datasets}
\end{table*}

\begin{table*}[h]
\centering
\normalsize
\setlength{\tabcolsep}{3pt}
\renewcommand{\arraystretch}{1.025}

\vskip -0.1in
\begin{tabular}{ C{1.4cm}  C{1.75cm} C{1.75cm} C{1.75cm}  C{1.75cm}}

 \hline
 \multirow{2}{*}{Methods}   & \multicolumn{2}{c}{Blurry (M=2K)} & \multicolumn{2}{c}{i-Blurry (M=2K)} \\
                            \cline{2-5} 
                            & ACC$\uparrow$ & FM$\downarrow$ & ACC$\uparrow$ & FM$\downarrow$ \\
 \hline
 
 CTN
 & 26.72$\pm$\small 0.67 & 21.51$\pm$\small 2.13
 & 33.75$\pm$\small 1.78 & 23.65$\pm$\small 2.31
 \\
 
 CCLL
 & 25.12$\pm$\small 2.07 & 22.34$\pm$\small 1.64
 & 28.18$\pm$\small 1.03 & 26.19$\pm$\small 1.35
 \\

 \hline
 DPCL
 & \textbf{47.58$\pm$\small 2.75} & \textbf{11.44$\pm$\small 2.64}
 & \textbf{50.22$\pm$\small 0.39} & \textbf{11.49$\pm$\small 2.54}
 \\
 \hline
\end{tabular}
\vskip -0.1in
\caption{Results of architecture-based methods on blurry~\cite{bang2021rainbow} and i-blurry~\cite{koh2022online} setups on CIFAR100. We measured averaged accuracy (ACC.) and forgetting measure(FM.) (\%) averaged by 5 different seeds.}
\label{table_architecture_setups}
\vskip -0.2in
\end{table*}

\subsection{Analysis on Weight Loss Landscape for Classifier}

From experiments in main paper, we analyzed the loss landscape on the function space of the first task data after training the fifth task. Through t-SNE visualization, we observed that our DPCL has relatively smaller loss values for the samples compared to the baselines, particularly showing a clear gap at class boundaries where the loss values are usually large. This indicates that our approach makes the function space smoother. To observe the sharpness of the classifier's weight loss landscape, we perturbed the weights of the trained classifier and examined how the loss values changed. In order to exclude the network's scaling invariance, we normalized the randomly sampled direction $\mathbf{d}$ from a Gaussian distribution using $\mathbf{d} \gets \frac{\mathbf{d}}{\lVert \mathbf{d} \rVert_{F}} \lVert \phi \rVert_F$~\cite{deng2021flattening}. The experiments were performed by averaging results over 5 random seeds.

\begin{figure*}[h]
\begin{center}
\centerline{\includegraphics[width=1\textwidth]{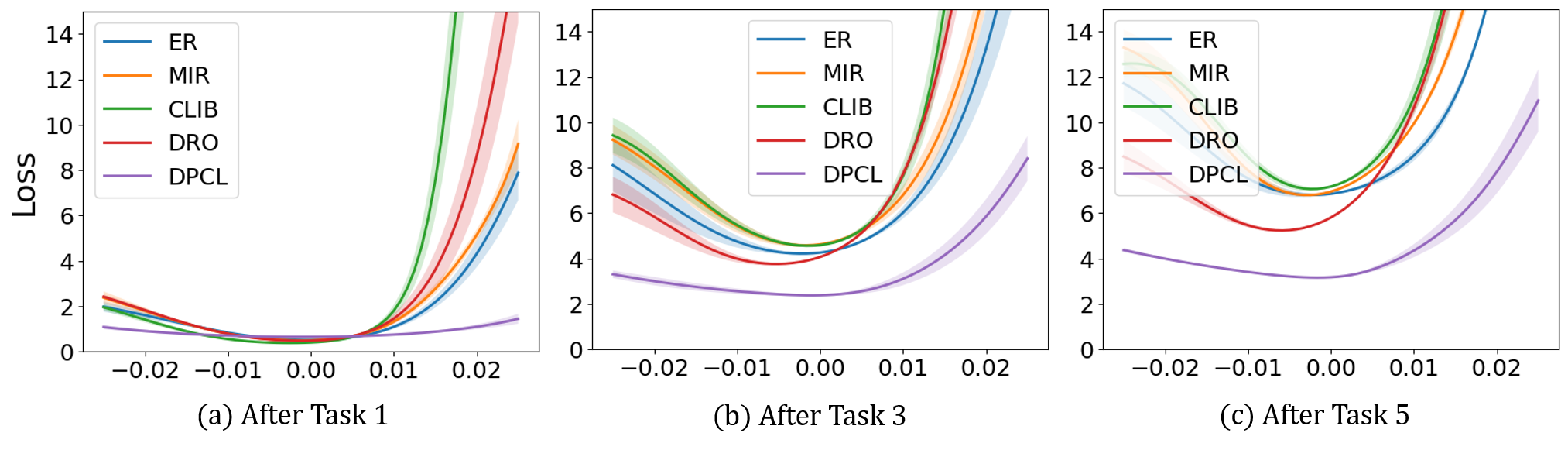}}
\caption{Weight loss landscape for data from first task after training for (a) first, (b) third, and (c) fifth task on CIFAR100 dataset. We randomly selected the direction for perturbation and averaged results from 5 random seeds. We can see that our DPCL has the flattest weight loss landscape for all cases and the lowest loss values at the origin.}
\label{figure5_weight_loss_landscape}
\end{center}
\vskip -0.25in
\end{figure*}

Figure~\ref{figure5_weight_loss_landscape} shows the loss landscapes after training the first, third, and fifth tasks on the first task data. Our DPCL exhibits the flattest loss landscape in all cases. Additionally, in Figure~\ref{figure5_weight_loss_landscape} (b) and (c), our DPCL consistently achieves the lowest loss across all regions. Note that among the baselines, ER exhibits the flattest loss landscape. This suggests that the existing methods are not necessarily related to flattening the classifier's weight loss landscape and can even deteriorate it. While DRO~\cite{wang2022improving} made the function space smoother in the experiments in the main paper by perturbing the input space, it does not exhibit favorable characteristics for the classifier's weight loss landscape.

\end{document}